\documentclass[10pt,twocolumn,letterpaper]{article}

\usepackage{iccv}
\usepackage{times}
\usepackage{epsfig}
\usepackage{graphicx}
\usepackage{amsmath}
\usepackage{amssymb}

\usepackage{times}
\usepackage{epsfig}
\usepackage{graphicx}
\usepackage{amsmath}
\usepackage{amssymb}
\usepackage{multirow}
\usepackage{tabularx}
\usepackage{booktabs}
\usepackage{colortbl}
\usepackage{arydshln}
\usepackage{subcaption}
\usepackage[dvipsnames]{xcolor}
\usepackage{tablefootnote}
\usepackage{enumitem}
\setlist[itemize]{leftmargin=*}

\usepackage{amsmath,amsfonts,bm}

\def\eqref#1{equation~\ref{#1}}
\def\1{\bm{1}}

\DeclareMathAlphabet{\mathsfit}{\encodingdefault}{\sfdefault}{m}{sl}
\SetMathAlphabet{\mathsfit}{bold}{\encodingdefault}{\sfdefault}{bx}{n}

\usepackage{xcolor}
\definecolor{asparagus}{rgb}{0.53, 0.66, 0.42}
\definecolor{bittersweet}{rgb}{1.0, 0.44, 0.37}

\newcommand{\specialcell}[2][c]{%
  \begin{tabular}[#1]{@{}c@{}}#2\end{tabular}}

\usepackage{tabu}

\usepackage[pagebackref=true,breaklinks=true,letterpaper=true,colorlinks,bookmarks=false]{hyperref}

\usepackage[font=small,labelfont=bf]{caption}  % set global caption size
\usepackage{makecell}
\usepackage{tabulary}
\definecolor{demphcolor}{RGB}{144,144,144}

\definecolor{mygray}{gray}{0.4}
\usepackage{pifont}% http://ctan.org/pkg/pifont for xmark and cmark
\definecolor{cmarkgreen}{RGB}{129,212,82}
\definecolor{xmarkred}{RGB}{219,58,37}
\newcommand{\cmark}{\ding{51}}%
\newcommand{\xmark}{\ding{55}}%
\newlength\savewidth

\makeatletter\renewcommand\paragraph{\@startsection{paragraph}{4}{\z@}
  {.5em \@plus1ex \@minus.2ex}{-.5em}{\normalfont\normalsize\bfseries}}\makeatother
\newcolumntype{C}[1]{>{\centering\arraybackslash}p{#1}}
\newcolumntype{R}[1]{>{\raggedleft\arraybackslash}p{#1}}
\newcolumntype{L}[1]{>{\raggedright\arraybackslash}p{#1}}

\iccvfinalcopy % *** Uncomment this line for the final submission

\def\iccvPaperID{9311} % *** Enter the ICCV Paper ID here

\ificcvfinal\fi

\begin{document}

%%%%%%%%% TITLE
\title{Adversarial VQA: A New Benchmark for \\ Evaluating the Robustness of VQA Models\\
\vspace{8pt}
\large{\url{adversarialvqa.github.io}}
\vspace{-8pt}
}
\author{Linjie Li$^1$, Jie Lei$^2$, Zhe Gan$^1$, Jingjing Liu$^3$\\ 
  $^1$Microsoft \quad $^2$UNC Chapel Hill \quad $^3$Tsinghua University \\
  \texttt{\small\{lindsey.li, zhe.gan\}@microsoft.com} \\
  \texttt{\small jielei@cs.unc.edu,}  
  \texttt{\small JJLiu@air.tsinghua.edu.cn}
  \\
  }

\maketitle
\ificcvfinal\thispagestyle{empty}\fi

%%%%%%%%% ABSTRACT
\begin{abstract}
Benefiting from large-scale pre-training, we have witnessed significant performance boost on the popular Visual Question Answering (VQA) task. Despite rapid progress, it remains unclear whether these state-of-the-art (SOTA) models are robust when encountering examples in the wild. To study this, we introduce Adversarial VQA, a new large-scale VQA benchmark, collected iteratively via an adversarial human-and-model-in-the-loop procedure. Through this new benchmark, we discover several interesting findings. ($i$) Surprisingly, we find that during dataset collection, non-expert annotators can easily  attack SOTA VQA models successfully. ($ii$) Both large-scale pre-trained models and adversarial training methods achieve far worse performance on the new benchmark than over standard VQA v2 dataset, revealing the fragility of these models while demonstrating the effectiveness of our adversarial dataset. ($iii$) When used for data augmentation, our dataset can effectively boost model performance on other robust VQA benchmarks. We hope our Adversarial VQA dataset can shed new light on robustness study in the community and serve as a valuable benchmark for future work. 
\end{abstract}

\begin{figure}
    \centering
    \includegraphics[width=.95\linewidth]{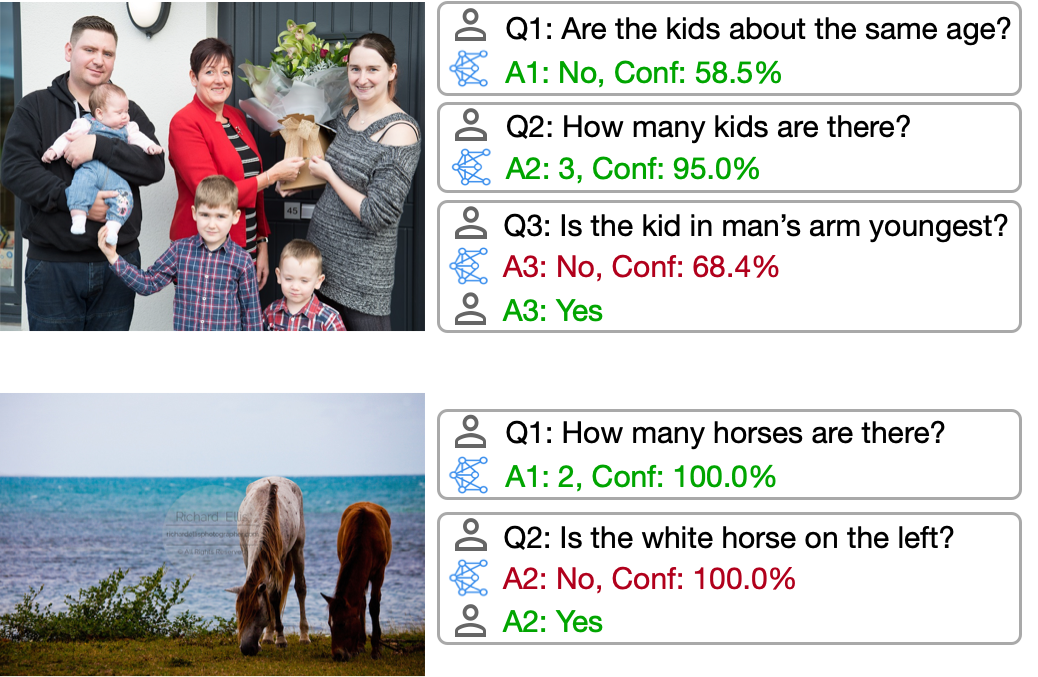}
    \vspace{-8pt}
    \caption{Illustration of data collection examples. The workers try to attack the VQA model for at most 5 times by asking \emph{hard} questions about the image, and succeeds at the last attempt. \textcolor{green}{Green} (\textcolor{red}{red}) indicates a correct (wrong) answer.}
    \label{fig:intro}
    \vspace{-10pt}
\end{figure}

%%%%%%%%% BODY TEXT
\section{Introduction}
\vspace{-1mm}
Visual Question Answering (VQA)~\cite{antol2015vqa} is a task where given an image and a question
about it, the model provides an open-ended answer.  A successful VQA system can be applied to real-life scenarios such as a chatbot that assists visually impaired people. In these applications, the VQA models are expected to handle diverse question types from recognition to reasoning, and answer questions faithfully based on the evidence in the image.

While model performance on the popular VQA dataset~\cite{goyal2017making} has been advanced  in recent years~\cite{antol2015vqa,jiang2018pythia,anderson2018bottom,yu2019mcan,chen2019uniter,tan2019lxmert,zhang2021vinvl}, with better visual representations~\cite{jiang2020defense,zhang2021vinvl}, more sophisticated model designs~\cite{gao2019dynamic,li2019relation}, large-scale pre-training~\cite{lu2019vilbert,su2019vl,cao2020behind,sun2021lightningdot,zhou2021uc2} and adversarial training~\cite{gan2020large}, today's VQA models are still far from being robust enough for practical use. There are some works studying the robustness of VQA models, such as their sensitivity to visual content manipulation~\cite{agarwal2020causal-vqa}, answer distribution shift~\cite{agrawal2018vqa-cp}, linguistic variations in input questions~\cite{shah2019vqa-rephrase}, and reasoning capabilities~\cite{gokhale2020vqa-lol,selvaraju2020vqa-introspect}.  However, current robust VQA benchmarks mostly suffer from three main limitations: ($i$) designed with heuristic rules~\cite{gokhale2020vqa-lol,agrawal2018vqa-cp,agarwal2020causal-vqa}; ($ii$) focused on a single type of robustness~\cite{selvaraju2020vqa-introspect,shah2019vqa-rephrase,gokhale2020vqa-lol}; ($iii$) based on VQA v2~\cite{goyal2017making} images (or questions), which state-of-the-art (SOTA) VQA models are trained on~\cite{gokhale2020vqa-lol,agrawal2018vqa-cp,agarwal2020causal-vqa,selvaraju2020vqa-introspect,shah2019vqa-rephrase}. The images~\cite{agarwal2020causal-vqa} or  questions~\cite{gokhale2020vqa-lol,hudson2019gqa} are often synthesized, not provided by human. 

In addition, previous data collection procedures on VQA benchmarks are often \emph{static}, meaning that the data samples in these datasets do not evolve, and model performance can saturate on the fixed dataset without good generalization. For example, model accuracy on VQA v2 has been improved from 50\%~\cite{antol2015vqa} to 76\%~\cite{zhang2021vinvl} since inception. Similarly, on robust VQA benchmarks, a recent study~\cite{li2020closer} has found that pre-trained models can greatly lift state of the art. Yet it remains unclear whether such high performance can be maintained when encountering examples in the wild.

To build an organically evolving benchmark, we introduce Adversarial VQA (AVQA), a new large-scale VQA dataset \emph{dynamically} collected with Human-And-Model-in-the-Loop Enabled Training (HAMLET)~\cite{xie2020adversarial}. AVQA is built on images from different domains, including web images from Conceptual Captions~\cite{sharma2018conceptual}, user-generated images from Fakeddit~\cite{fakeddit}, and movie images from VCR~\cite{zellers2019vcr}. Our data collection is iterative and can be perpetually going. We first ask human annotators to create examples that current best models cannot answer correctly (Figure~\ref{fig:intro}). These newly annotated examples expose the model's weaknesses, and are added to the training data for training a stronger model. The re-trained model is subjected to the same process, and the collection can iterate for several rounds. After each round, we train a new model and set aside a new test set. In this way, not only is the resultant dataset more challenging than existing benchmarks, but this process also yields a ``moving post” target for VQA systems, rather than a static benchmark that will eventually saturate.

With this new benchmark, we present a thorough quantitative evaluation on the robustness of VQA models along multiple dimensions. First, we provide the first study on the vulnerability of VQA models when under adversarial attacks by human. 
Second, we benchmark several SOTA VQA models on the proposed dataset to reveal the fragility of VQA models. We observe a significant and universal performance drop when compared to VQA v2 and other robust VQA benchmarks, which corroborates our belief that existing VQA models are not robust enough. Meanwhile, this also demonstrates the transferability of these adversarial examples -- data samples collected using one set of models are also challenging for other models. Third, as our annotators can ask different types of questions for different types of robustness, our analyses show that SOTA models suffer across various questions types, especially counting and reasoning. 

Our main contributions are summarized as follows. ($i$) For better evaluation of VQA model robustness, we introduce a new VQA benchmark dynamically collected with a Human-and-Model-in-the-Loop procedure. 
($ii$) Despite rapid advances on VQA v2 and robust VQA benchmarks, the evaluation on our new dataset shows that SOTA models are far from being robust. In fact, they are extremely vulnerable when attacked by human annotators, who can succeed within 2 trials on average. ($iii$) We provide a thorough analysis to share insights on the shortcomings of current models as well as comparison with other robust VQA benchmarks.

%-------------------------------------------------------------------------
\vspace{-5pt}
\section{Related Work}
\vspace{-1mm}
\noindent\textbf{Robust VQA Benchmarks}\, 
There has been a growing interest in building new benchmarks to study the robustness of VQA models. VQA-CP~\cite{agrawal2018vqa-cp}, the first robust VQA benchmark constructed via reshuffling examples in VQA v2~\cite{goyal2017making}, is proposed
to evaluate question-oriented language bias in VQA models. GQA-OOD~\cite{kervadec2020gqa-ood} improves from VQA-CP, and proposes to evaluate the performance differences between in-distribution and out-of-distribution split. 
Besides language bias, VQA-Rephrasings~\cite{shah2019vqa-rephrase} exposes the brittleness of VQA models to linguistic variations in questions by collecting human-written rephrasings of VQA v2 questions. 
Causal VQA~\cite{agarwal2020causal-vqa} studies robustness against semantic image manipulations, and tests for prediction consistency to questions on clean images and corresponding edited images. Further studies investigate robustness against reasoning. For instance, \cite{selvaraju2020vqa-introspect} collects perception-related sub-questions per question for a new reasoning split of VQA dataset.
\cite{gokhale2020vqa-lol} tests model's ability to logical reasoning through logical compositions of yes/no questions in VQA v2.
GQA~\cite{hudson2019gqa} provides large-scale rule-based questions from ground-truth scene graphs, that can test VQA model's ability on positional reasoning and relational reasoning.

Despite the continuous efforts in evaluating robustness of VQA models, these works mostly focus on a single type of robustness, and are based on the original VQA v2 dataset via either another round of question collection given the existing VQA examples, or automatic transformation or manipulation of current examples. In comparison, we use different image sources, and collect a new challenging VQA benchmark by allowing human annotators to directly attack current state-of-the-art VQA models.

\begin{figure*}
    \centering
    \includegraphics[width=0.86\linewidth]{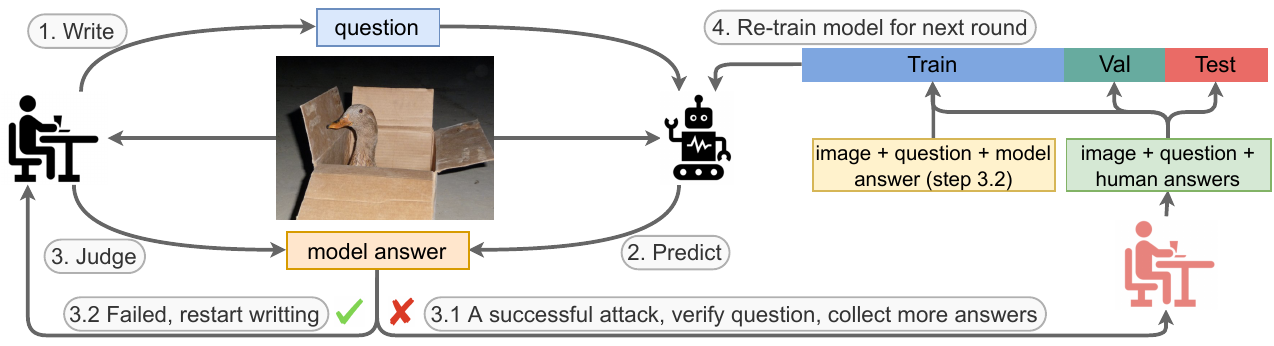}
    \vspace{-6pt}
    \caption{Overview of our adversarial data collection process, for a single round. The process can be considered as a game played by two parties, a human annotator and a well-trained model. Given an image, the annotator tries to attack the model by writing a tricky question (\textit{step 1}), the model then predicts an answer to the question (\textit{step 2}). Next, the human annotator judges the correctness of the model answer (\textit{step 3}). If the model answer is judged as ``definitely 
   wrong'' {\color{xmarkred}\xmark}, meaning the attack is successful, then we verify the question and collect more answers for it (\textit{step 3.1}). Otherwise, the attack is failed, the annotator needs to write another question to attack the model (\textit{step 3.2}). The \textit{val} and \textit{test} splits contain only successfully attacked questions, while \textit{train} split contains also the failed questions.}
    \label{fig:collection_process}
    \vspace{-8pt}
\end{figure*}

\vspace{2pt}
\noindent\textbf{Model-in-the-Loop Data Collection}\, 
Dataset collection with a \textit{model-in-the-loop} setting has received increasing attention in recent years in the NLP community. In this setting, models are used in the collection process to identify wrongly predicted, thus more challenging examples. These models are used either as a post-processing filter~\cite{zellers2018swag,bartolo2020beat} or directly during annotation~\cite{yang2017mastering,nie2019adversarial,bartolo2020beat}. 
In ANLI~\cite{nie2019adversarial}, the \textit{model-in-the-loop} strategy is extended to a \textit{Human-And-Model-in-the-Loop Enabled Training} (HAMLET) setting, where the data collection happens in multiple rounds, and in each round, the models are updated to stronger versions by training with examples collected from previous rounds.
The goal of ANLI is to create a natural language inference (NLI) dataset that can grow along with the rapid advance of model capabilities~\cite{devlin2018bert,liu2019roberta,yang2019xlnet,lan2019albert}.
In contrast to static datasets that will eventually saturate as models become stronger, datasets created with the HAMLET procedure are dynamic -- if the test set saturates with a more powerful model, one can use this more powerful model to assist the collection of a new set of difficult examples, leading to a never-ending challenge for the community.
Meanwhile, the adversarial nature of the HAMLET procedure also helps to identify the weaknesses and vulnerabilities of existing models, and the biases or annotation artifacts~\cite{gururangan2018annotation,poliak-etal-2018-hypothesis,lei2020more} in existing datasets~\cite{bowman2015large,williams2017broad,lei2020more}.
Beyond its application to the NLI task, the HAMLET procedure is also proved to be useful in collecting more challenging examples for the video-and-language future prediction task~\cite{lei2020more}. 

\begin{table*}[t!]
    \centering
\small
\resizebox{\textwidth}{!}{
    \small
    \begin{tabu}{lcccccccc}
    \hline
    \multirow{2}{*}{Dataset} & \multirow{2}{*}{Image Source} & \multirow{2}{*}{\#Image} & \multirow{2}{*}{IsCollected} &\#IQ & Model error rate (\%)  & \#Tries & Time (sec.) & Data Split\\
     \cmidrule(lr){5-5} \cmidrule(lr){6-6} \cmidrule(lr){7-7} \cmidrule(lr){8-8}\cmidrule(lr){9-9}
    & & & & Total/Verified & Total/Verified  & Mean/Median & per verified ex. & Train/Val/Test\\
    \hline
    \hline
    
    \multicolumn{3}{l}{\textbf{Previous Robust VQA Datsets}}\\
    \hline
    \hline
    VQA-Reph.& \multirow{5}{*}{COCO}& - & \cmark & 162K/- & - & - & - & -/162K/-\\
    VQA-Intro. & & - & \cmark & 238K/- & - & - & - & 222K/-/93K\\
    VQA-LOL Comp.& & - & \xmark  & 1.25M/- & - & - & - & 916M/43K/291K\\
    VQA-LOL Supp. &  & - & \xmark & 2.55M/- & - & - & - & 1.9M/9k/669K\\
    VQA-CP v2&  & -  & \xmark & -/- & - & - & - & 438K/-/220K\\
    \hline
    IV-VQA & \multirow{2}{*}{COCO$^\dagger$} & 357K & \xmark & 376K/- & - & - & - & 257K/11.6K/108K\\
    CV-VQA &  & 18.0K & \xmark & 12.7K/- & - & - & - & 8.5K/0.4K/3.7K\\
    \hline
    \hline
    \multicolumn{3}{l}{\textbf{Ours}}\\
    \hline
    \hline
    % All 13740 images are split into 8522 Train, 1374 Val and 4122 Test.
    % Train Verified 25365 QAs, Unverified 28223 QAs
    % Val 3319 QAs
    % Test 9983 QAs
    % 32745 QA valid, meaning verified model failed
    
    % Loaded 49489 data
    % Total # questions: 93131
    % Total verified examples: 45565
    % Average time cost: 71.0395710523432
    % Error rate: 0.48925706800098784
    % Mean Tries 1.6324371776582904
    % Median Tries 1
    % R1 & CC~\cite{sharma2018conceptual} & 13,740 & 93,131/45,565 & 48.93\%/35.16\% & 1.63/1 & 71.04 &  53,588/3,319/9,983\\ 
    R1 & CC & 13.7K & \cmark & 93.1K/45.6K & 48.9/35.2 & 1.6/1 & 71.0 &  53.6K/3.3K/10.0K\\ 
    % All 13131 images are split into 8717 Train, 1313 Val and 3940 Test.
    % Train Verified 23469 QAs, Unverified 19,340
    % val 2716 QAs
    % Test 8287 QAs
    % All 41,972 + 37,835

    % Loaded 41972 data
    % Total # questions: 70401
    % Total verified examples: 39472
    % Average time cost: 54.16573467267942
    % Error rate: 0.5606738540645729
    % Mean Tries 1.5098804215646535
    % Median Tries 1
    % R2 & CC & 13,131 & 70,401/37,835 & 56.07\%/53.74\% &1.51/1 & 54.17 & 42,809/2,716/8,287\\ 
    R2 & CC & 13.1K & \cmark & 70.4K/37.8K & 56.1/49.0 &1.5/1 & 54.2 & 42.8K/2.7K/8.3K\\
    % All 7765 images are split into 4964 Train, 776 Val and 2330 Test.
    % Writing 13655 QAs
    % Writing 1839 QAs
    % Writing 5480 QAs
    % Loaded 27532 data
    % In total, 15697 QAs 
    
    % Total # questions: 50215
    % Total verified examples: 25017
    % Average time cost: 57.30335679737783
    % Error rate: 0.4981977496763915
    % Mean Tries 1.6075868409481553
    % Median Tries 1
    % R3 & Various & 7.8K & $\checkmark$ & 50.2K/25.0K & 49.8\%/36.3\% & 1.6/1 & 57.3 & 13.7K/1.8K/5.5K\\ 
    % R3 & Various & 11.1K & $\checkmark$ & 60.4K/30.0K & 49.8\%/36.3\% & 1.6/1 & 57.3 & 35.3K/2.2K/6.6K\\ 

    % Total # questions: 79525
    % Total verified examples: 40271
    % Average time cost: 57.33891003451657
    % Error rate: 0.5063942156554543
    % Mean Tries 1.5988925032902088
    % Median Tries 1
    R3 & Various & 11.1K & \cmark & 79.5K/40.3K & 50.7/34.4 & 1.6/1 & 57.3 & 45.9K/2.7K/8.1K\\ 
    \hline
    % AVQA & Various & 34.6K & $\checkmark$& 213.7K/108.4K & 50.7\%/40.0\% & 1.6/1 & 61.6 & 110.1K/7.8K/23.8K\\ 
    % AVQA & Various & 37.9K & $\checkmark$& 223.9K/113.4K & 50.7\%/40.0\% & 1.6/1 & 61.6 & 131.7K/8.2K/24.9K\\ 
    AVQA & Various & 37.9K & \cmark & 243.0K/123.7K & 50.9/38.1 & 1.6/1 & 61.3 & 142.1K/8.7K/26.4K\\ 
    \hline
    \end{tabu}
}
\vspace{-8pt}
    \caption{\small{Data statistics. `Model error rate' is the percentage of examples that the model gets wrong; `Verified' is the questions with 10 answer annotations. Images for R3 are from various domains: Conceptual Captions (CC)~\cite{sharma2018conceptual}, VCR~\cite{zellers2019vcr} and Fakeddit~\cite{fakeddit}. We compare our dataset against previous robust VQA datasets, based on COCO~\cite{chen2015microsoft} images. For number of image-question pairs (\#IQ) and images (\#Image), we only report the number of new examples generated/collected in each dataset. $\dagger$ indicates that the images are not natural, but edited. `IsCollected' indicates whether the data is collected via crowdsourcing.
    }}
    \label{tab:data_stats}
    \vspace{-3mm}
\end{table*}
%-------------------------------------------------------------------------
\vspace{-2pt}
\section{Adversarial VQA Dataset}
\vspace{-1mm}
In this section, we introduce the AVQA dataset in detail. Sec.~\ref{sec:pipeline} explains the data collection pipeline. Sec.~\ref{sec:statistics} and Sec.~\ref{sec:comparison} present data statistics and the comparison with other datasets.

\subsection{Data Collection Pipeline}~\label{sec:pipeline}
The HAMLET data collection procedure can be considered as a game played by two parties: a human annotator and a well-trained model.
The human annotator competes against the model as an adversary and tries to design adversarial examples to identify its vulnerabilities. After collecting enough examples, the model augments its training with the collected data to defend similar attacks. For VQA, we define the adversarial example as an adversarial question on a natural image that the model answers incorrectly. 

As shown in Figure~\ref{fig:collection_process},
given an image, the human annotator tries to write a \emph{tricky} question that the VQA model may fail. Once the question is submitted, an online model prediction will be displayed immediately to the workers. The model answer is then judged by the same annotator as either ``definitely correct", ``definitely wrong", or ``not sure". If the model prediction is ``definitely wrong", then the attack is successful, and we further ask the annotator to provide a correct answer.
Otherwise, the annotator needs to write another question until the model predicts a wrong answer, or the number of tries exceeds a threshold (5 tries). To avoid obviously invalid questions caused by the annotator taking shortcuts (\emph{e.g.}, untruthful judgement on model predictions, questions irrelevant to the image content), we also launch an answer annotation task. 
Successfully attacked questions are provided to 9 other annotators to collect extra answers, as well as their confidence level (``confident", ``maybe" and ``not confident") of their answer.
The questions that receive less than 6 ``confident" answers and have no agreement in answers among 10 annotators are removed during post-processing. In the end, each image is presented to 3 workers for question collection, and each image-question pair is shown to 10 annotators for answer collection.

This procedure can be continuously deployed for multiple rounds. At each round, we strengthen the models as we re-train them with extra data collected from previous rounds. This ``dynamic" evolution of attacked models allows the collection of ``harder" questions in the later rounds. In our setup, we launch the data collection for 3 rounds on Amazon Mechanical Turk. 
However, this data collection can be a never-ending process, as we can always replace the attacked model with a stronger model trained on newly collected data or better architectures developed in the future.

\vspace{2pt}
\noindent\textbf{Round 1 (R1)}\,
For the first round, we employ VQA models trained on examples from VQA v2~\cite{goyal2017making} and VGQA~\cite{krishna2017visual} as our starting point.
To avoid the collected questions overfitting to the vulnerabilities of a single model or a single architecture, for each user question, we randomly sample one model from LXMERT~\cite{tan2019lxmert}, UNITER-B~\cite{chen2019uniter} and UNITER-L~\cite{chen2019uniter} as the attacked model to generate the answer.
We choose LXMERT and UNITER as representatives of two-stream and single-stream pre-trained V+L models, due to their strong performance on VQA v2. 
We use images sampled from Conceptual Captions~\cite{sharma2018conceptual} for annotation. In total, we collected 38.7K verified\footnote{Verified questions are all successfully attacked questions.} questions and 28.2K unverfied questions over 13.7K images, and split the verified examples into 60\%/10\%/30\% for train/val/test splits. All unverified examples are also added to the training split.

\vspace{2pt}
\noindent\textbf{Round 2 (R2)}\, For the second round, we re-train our models with questions from VQA v2, VGQA and R1's train split, and select the best model checkpoints of LXMERT, UNITER-B and UNITER-L based on R1's val set. Similarly, we randomly sample one model at a time for the workers to attack. A new set of non-overlapping Conceptual Captions images are used. In total, we collected 23.5K verified questions and 19.3K unverified question over 13.1K images, and split the data in a similar manner to R1. 

\vspace{2pt}
\noindent\textbf{Round 3 (R3)}\, For the third round, we include more diverse images from different domains: ($i$) web images from Conceptual Captions~\cite{sharma2018conceptual}; ($ii$) user-generated images from Fakeddit~\cite{fakeddit}; and ($iii$) movie frame images from VCR~\cite{zellers2019vcr}. The attacked model is still randomly sampled from LXMERT, UNITER-B and UNITER-L, but we add the training set from R1 and R2 to the training data.

\vspace{2pt}
\noindent\textbf{Summary}\,
Finally, combining data collected in R1, R2 and R3 produces our proposed AVQA dataset. In the end, we collected 243.0K questions over 37.9K images, with 142.1K/8.7K/26.4K images in the train/val/test split.

\subsection{Data Statistics}~\label{sec:statistics}\, The data statistics of the new dataset are summarized in Table~\ref{tab:data_stats}. The number of examples we collected per image varies per round, starting with approximately 6.8 questions/image for R1, to around 5.4 for R2 and 7.2 for R3. 
Under the same image domain for R1 and R2, we suspect that the annotators learn to identify model vulnerabilities more rapidly than the models learn to defend itself from the adversarial examples. We provide analyses in Sec.~\ref{sec:eval} and~\ref{sec:q_type} for further investigation. 
On the one hand, the annotators are getting better at identifying vulnerabilities of these models. Analyses of question types per round in Sec.~\ref{sec:q_type} show that the workers tend to ask more questions in certain categories, such as ``count", ``OCR" and ``commonsense reasoning", that the model is more likely to fail. On the other hand, although the attacked model is strengthened through data augmentation, the model does not seem to learn from the adversarial examples effectively.

For each round, we report the model error rate, both on verified and all examples. The model error rate reported under ``Total" captures the percentage of examples where the writer disagrees with the model's answer during question collection, but where we are not yet sure that the example is correct. The verified model error rate is the percentage of model errors from examples that we further collected 9 additional answers from other workers. We observe an increase in model error rate from R1 to R2. Assuming constant image domain difficulty in R1 and R2, the higher model rate suggests that the models in the later rounds are not significantly stronger, or the annotators are getting better at fooling the state-of-the-art models. In R3, where we included images from more diverse domains, the model error rate decreases from 49.0\% to 34.4\%. We suspect it is because the movie images from VCR are mostly human-centric, which is commonly observed in COCO. 

We also report the average number of attempts (``\#Tries" in Table~\ref{tab:data_stats}) that a worker needed to complete the annotation process for each image, \emph{i.e.}, to successfully attack the model or exceed the limits on number of tries. Surprisingly, although the VQA models used in the later rounds are trained with more data, the number of tries needed to successfully attack them does not increase. On average, it takes less than 2 tries to successfully attack a VQA model. Similarly, the average time needed per successful attack decreases by 15 seconds as data collection progresses. 

\begin{table*}[hbt!]
    \centering
\small

    \small
    \begin{tabu}{llccccc|c}
    \hline
    \multirow{2}{*}{Model} & \multirow{2}{*}{Training Data} & R1 & R2 & R3 & AVQA & VQA v2 & $\Delta$(v2, AVQA)\\
    \cmidrule(lr){3-3} \cmidrule(lr){4-4} \cmidrule(lr){5-5} \cmidrule(lr){6-6} \cmidrule(lr){7-7} \cmidrule(lr){8-8}
    & & val/test & val/test & val/test &  val/test & test-dev & test-dev, test\\
    \hline
    \multirow{2}{*}{BUTD} & VQA v2 +VGQA & 20.80/19.28& 18.77/18.85 & 20.63/21.10 & 20.12/19.71 & 67.60 & 47.89 \\
    %  & +R1 & 20.27/20.27& 19.53/20.14 & 21.34/21.82 & 20.26/20.58 & 67.37 & 46.79\\
    %  & +R1+R2 & 24.41/21.82 & 22.28/21.80& 21.47/20.94 & 23.00/21.61 & 67.44 & 45.83\\
     & ALL & 24.96/22.11 & 22.62/22.78 & 23.92/23.61 & 23.91/22.78 & 67.52 & \textbf{44.74}\\
    \hline
    \multirow{4}{*}{UNITER-B} & VQA v2 +VGQA &\underline{20.60/17.91} & 17.86/18.55  & 20.71/20.17& 19.79/18.81 &72.70 & 53.89\\
     & +R1 & 26.03/22.94 & \underline{17.30/17.36} & 20.56/20.61 & 21.62/20.47 & 72.98 & 52.51\\
     & +R1+R2 & 26.60/24.76 & 23.21/23.86 & \underline{19.26/18.73} & 23.26/22.62 & 72.75 & 50.13\\
     & ALL & 26.85/24.93 & 23.38/23.92 & 24.48/23.27 & 25.04/24.10 & 72.66 & 48.56\\
    \hline
    \multirow{4}{*}{UNITER-L} & VQA v2 +VGQA & \underline{25.04/23.72}&  17.82/17.49 & 19.63/19.77& 21.12/20.55 & 73.82 & 53.27\\
     & +R1 & 29.31/26.63 & \underline{19.34/18.66} & 19.78/18.99 & 23.25/21.78 & 73.89 & 52.11\\
     & +R1+R2 & 30.13/28.15 & 23.11/23.54 & \underline{17.35/17.05} & 23.97/23.29 & 73.77 & 50.48\\
     & ALL & \textbf{30.80}/\textbf{28.45} & 22.95/23.11 & 24.08/21.97 &  \textbf{26.27}/\textbf{24.78}& \textbf{74.15} & 49.37\\
    \hline
    \multirow{4}{*}{LXMERT} & VQA v2 +VGQA & \underline{19.76/18.15}& 18.98/18.79 & 21.08/21.27 & 19.93/19.31 & 72.31 &53.00\\
     & +R1 & 23.89/22.65 & \underline{19.01/17.91} & 21.64/21.42& 21.68/20.78 & 72.51 & 51.73\\
     & +R1+R2 & 26.76/24.86 & 23.28/\textbf{24.11} & \underline{19.39/19.57} & 23.38/23.00 & 72.61 & 49.61\\
     & ALL & 26.35/24.55 & \textbf{23.84}/24.02 & \textbf{25.27}/\textbf{23.71}& 25.24/24.13 & 72.42 & 48.29\\
    \hline
    \end{tabu}
\vspace{-8pt}
    \caption{\small{Model performance of various models under different settings. AVQA / ALL refers to R1+R2+R3 / VQA v2+VGQA+AVQA.}}
    \label{tab:model_eval}
    \vspace{-3mm}
\end{table*}

\subsection{Comparison with Other Datasets}~\label{sec:comparison}
Our Adversarial VQA dataset sets a new benchmark for evaluating the robustness of VQA models. It improves upon existing robust VQA benchmarks in several ways. First, the dataset by design is more difficult than previous datasets. During collection, we do not constrain the worker to ask questions that only fall into a single robustness type (Sec.~\ref{sec:q_type}). 
As a result, our dataset is helpful in defending model robustness against several robust VQA benchmarks (Sec.~\ref{sec:robust_eval}). 
Second, most robust VQA datasets are based on VQA v2 validation set, which state-of-the-art models use for training or hyper-parameter tuning. Thus, it is difficult to analyze the robustness of the best-performing models due to this data leakage. 
Our dataset is built on non-overlapping images from diverse domains, which naturally resolves it.
Lastly, our dataset is composed of human-written questions on natural images, rather than rule-based questions in ~\cite{gokhale2020vqa-lol,hudson2019gqa} or manipulated images in~\cite{agarwal2020causal-vqa}. A detailed comparison on data statistics is provided in Table~\ref{tab:data_stats}.

Our work is inspired by ANLI~\cite{nie2019adversarial}. While ANLI focuses on the pure text task of natural language inference, our work targets at the multi-modal task of visual question answering. 
However, due to the open-ended nature of VQA problem, the construction of AVQA is more challenging. 
Instead of giving the worker a target label when collecting adversarial questions, we first ask the worker to judge whether the model prediction is correct, then provide a ground-truth answer. Our verification process is also different from ANLI. In order to evaluate model performance under the same criteria as VQA v2~\cite{goyal2017making}, we collect 10 answers from workers in total. Unlike the observations on ANLI, where the adversarial robustness of NLI models can be improved in a large extent through data augmentation of ANLI, our analysis on AVQA in Sec.~\ref{sec:res} will show that it is more difficult to defend against adversarial attacks for VQA models.

%------------------------------------------------------------------------
\vspace{-2pt}
\section{Experiments and Analysis}~\label{sec:res}
% \vspace{-1mm}
In this section, we conduct extensive experiments to study the AVQA dataset. Specifically, Sec.~\ref{sec:eval} and Sec.~\ref{sec:deep_ana} evaluate different model architectures with different modality inputs on AVQA; Sec.~\ref{sec:robust_eval} examines how AVQA can help over other popular robust VQA benchmarks; Sec.~\ref{sec:q_type} explores the question types that can fool the models; and Sec.~\ref{sec:attack_methods} compares our data collection with automatic adversarial attack methods both qualitatively and quantitatively. 

\subsection{Model Evaluation}~\label{sec:eval} 
Table~\ref{tab:model_eval} reports the main results. In addition to UNITER-B, UNITER-L~\cite{chen2019uniter} and LXMERT~\cite{tan2019lxmert}, we also include BUTD~\cite{anderson2018bottom} as an example of task-specific model with different model architecture, prior to the large-scale pre-training era.  We show performance
on the AVQA test sets per round, the total AVQA test set, and VQA v2 test-dev set. Our key observations are summarized as follows.

\vspace{2pt}
\noindent\textbf{\emph{O1: Adversarial examples are transferrable across models.}} Both LXMERT and UNITER are variants of Transformer~\cite{transformer} architecture. We use BUTD as an example to investigate whether the adversarial examples are transferrable among the three models. 
The $\sim$20 performance of BUTD (trained on VQA v2+VGQA) on test set of each round indicates that workers did not find vulnerabilities specific to a single model architecture, but generally applicable ones across different model architectures.

\vspace{2pt}
\noindent\textbf{\emph{O2: The difficulty level of rounds does not decrease.}} Under the same training data, we observe that the model achieves comparable or even lower performance on later rounds. As aforementioned in data statistics, the increased model error rates and the decreased average tries annotators needed suggest that the later rounds contain more difficult examples.

\begin{table}[t!]
    \centering
\small
\begin{tabu}{l|ccc}
    \hline
    Data  & R1 &  R2 & R3 \\
    \hline
    Verified & 25.63 & 22.84 & 23.63 \\
    Combined & 26.85 & 22.82 &  24.38 \\
    \hline
\end{tabu}
\vspace{-8pt}
    \caption{\small{Comparison of verified and combined data. Results are reported on val split from UNITER-B trained on training data of each round, VQA v2 and VGQA.}}
    \label{tab:verfied_combined}
    \vspace{-6pt}
\end{table}

\vspace{2pt}
\noindent\textbf{\emph{O3: Training with more rounds help defend robustness...} } Generally, our results indicate that training on more rounds improves model performance. 
% \begin{table*}[!t]
% \begin{minipage}{.6\textwidth}
%     \centering
%     \small

%  \begin{tabu}{lcccc}
%     \hline
%      \multirow{2}{*}{Training Data} & R1 & R2 & R3  & VQA v2 \\
%     \cmidrule(lr){2-2} \cmidrule(lr){3-3} \cmidrule(lr){4-4} \cmidrule(lr){5-5} 
%     &  val & val & val  & val \\
%     \hline
%     VQA v2+VG & 18.93 & 17.07  & \\
%     All & & & & \\
%     \hline
%     VQA v2+VG$^L$ & 18.93 & 17.07  & \\
%     All$^L$ & & & & \\
    
%     \hline
%     \end{tabu}

% \vspace{-6pt}
%      \caption{\small{Language-only Model Performance. The models are trained by zeroing out image features.}}
%     \label{tab:Lonly_model_eval}

% \end{minipage}\hfill
% \begin{minipage}{.34\textwidth}
%     \centering
%     \small
% \resizebox{.99\textwidth}{!}{
% \begin{tabu}{l|cc}
%     \hline
%     \multirow{2}{*}{Model}  & AVQA &  VQA-CP v2 \\
%     \cmidrule(lr){2-2} \cmidrule(lr){3-3} 
%     & val &  test\\
%     \hline
%     BUTD & &  40.62 (38.82~\cite{teney2020value})\\
%     %  & +R1 & 21.11/20.41 & & & \\
%     %  & +R1+R2 & & & & \\
%     %  & +R1+R2+R3 & & & & \\
%     +~\cite{teney2020value}  &  &  44.XX (55.15~\cite{teney2020value})\\
%     \hline
%     UNITER-B &  & 47.02 (46.93~\cite{li2020closer})\\
%     %  & +R1 & 20.12/19.74 & & & \\
%     %  & +R1+R2 & & & & \\
%     %  & +R1+R2+R3 & & & & \\
%     +~\cite{teney2020value}  &  & 47.12\\
%     \hline
%     \end{tabu}
% }
% \vspace{-6pt}
% \caption{\small{Model Performance with a VQA-CP baseline from~\cite{teney2020value}.}}
%     \label{tab:vqa-cp}
% \end{minipage}
% \vspace{-3mm}
% \end{table*}

\begin{table}[!t]
% \tablestyle{5pt}{1.1} 
\def\w{20pt} 
\centering

\begin{subtable}{0.9\linewidth}
% \begin{subtable}{0.8\linewidth}
    \centering
    \small
\resizebox{\linewidth}{!}{
 \begin{tabu}{lccccc}
    \hline
     Training & Lang.  & R1 & R2 & R3  & VQA v2 \\
    \cmidrule(lr){3-3} \cmidrule(lr){4-4} \cmidrule(lr){5-5} \cmidrule(lr){6-6} 
    Data & Only & test & test & test  & test-dev \\
    \hline
    VQA v2+VG & \xmark & 17.91 & 18.55 & 20.17 & \textbf{72.70}\\
    AVQA-only & \xmark& \textbf{25.66} & \textbf{24.91} & \textbf{24.75} & 59.99\\
    ALL & \xmark& 24.93 & 23.92 & 23.27 & 72.66\\
    \hline
    VQA v2+VG & \cmark& 17.82 & 17.03 & 21.32 & 45.81\\
    AVQA-only & \cmark& 20.37 & 21.49 & 22.89 & 38.21\\
    ALL & \cmark& 19.75 & 20.75 & 22.81 & 46.23\\
    
    % \hline
    %  \multirow{2}{*}{Training Data} & R1 & R2 & R3  & VQA v2 \\
    % \cmidrule(lr){2-2} \cmidrule(lr){3-3} \cmidrule(lr){4-4} \cmidrule(lr){5-5} 
    % &  test & test & test  & test-dev \\
    % \hline
    % VQA v2+VG & 17.91 & 18.55 & 20.20 & \textbf{72.70}\\
    % AVQA-only & \textbf{25.66} & \textbf{24.91} & \textbf{25.06} & 59.99\\
    % All & 24.93 & 23.92 & 23.02 & 72.66\\
    % \hline
    % VQA v2+VG$^L$ & 17.82 & 17.03 & 21.19 & 45.81\\
    % AVQA-only$^L$ & 20.37 & 21.49 & 22.92 & 38.21\\
    % All$^L$ & 19.75 & 20.75 & 22.31 & 46.23\\    
    
    \hline
    \end{tabu}
}
     \caption{\small{Language-only model performance.}
    %   (indicated by $L$). The models are trained by zeroing out image features.
     }
    \label{tab:Lonly_model_eval}
\end{subtable}
\vspace{2pt}

\begin{subtable}{1.0\linewidth}
    \centering
    \small
\resizebox{0.66\linewidth}{!}{
\begin{tabu}{l|cc}
    \hline
    \multirow{2}{*}{Model}  & AVQA &  VQA-CP v2 \\
    \cmidrule(lr){2-2} \cmidrule(lr){3-3} 
    & val &  test\\
    \hline
    BUTD & \textbf{23.91} &  40.62 (38.82~\cite{teney2020value})\\
    +~\cite{teney2020value}  & 23.79 &  \textbf{43.96} \\
    \hline
    UNITER-B & \textbf{25.04} & 47.02 (46.93~\cite{li2020closer})\\
    +~\cite{teney2020value}  & 24.70 & \textbf{47.12}\\
    \hline
    \end{tabu}
}

\caption{\small{Model performance with a VQA-CP baseline from~\cite{teney2020value}.}}
    \label{tab:vqa-cp}
\end{subtable}

\vspace{-6pt}
\caption{\small{Analysis on language bias.}}
\label{tab:l-bias}
\vspace{-6pt}
\end{table}

\vspace{2pt}
\noindent\textbf{\emph{...but data augmentation alone is not effective.}} To investigate how much improvements are from adversarial examples, we show comparison of UNITER-B results on verified and combined data in Table~\ref{tab:verfied_combined}. In addition to verified data, the combined data include examples that the worker thinks the model has answered correctly. Even with almost doubled data size, results on combined data are not significantly better. This implies that simply training on more examples that the model correctly answers can hardly help the model be robust to adversarial attacks. 

\vspace{2pt}
\noindent\textbf{\emph{O4: Large model does not possess a clear advantage.}} Although outperforming UNITER-B and LXMERT on R1, UNITER-L does not show a clear advantage over R2 and R3. Overall, these three models achieve similar performance across rounds and on AVQA. When trained with ``ALL'' data, the performance gain from UNITER-L over BUTD is only +2.00 on AVQA, even though UNITER-L is pre-trained with extensive amount of image-text pairs.

\subsection{Key Factor Analysis}\label{sec:deep_ana}
We dive deeper into the key factors behind the low performance of state-of-the-art models on AVQA, and try to answer the following questions.

\vspace{2pt}
\noindent\textbf{\emph{Q1: Is the language in AVQA biased?}} 
Starting from VQA-CP~\cite{agrawal2018vqa-cp}, concerns have been raised about the propensity of models to pick up on spurious artifacts that are present just in the co-occurrence of question-answer pairs, without actually paying attention to the image content. We compare full models trained with both images and questions to models trained only on questions by zeroing out image features in Table~\ref{tab:Lonly_model_eval}. The results show that language-only models perform poorly on AVQA, and similarly on VQA v2.  Language-only model performance decreases over rounds for AVQA. However, UNITER-B is not much better than language-only on AVQA. Obviously, without manual intervention, some bias remains in how annotators phrase questions. For example, there might be more counting questions with answers other than 2, which is the majority answer in VQA v2. Therefore, models trained on AVQA only performs slightly higher for both UNITER-B and Language-only model. However, we also observe the significant drop in VQA v2 performance is out of proportion to the slight performance improvement on AVQA.
% \begin{table*}[hbt!]
%     \centering
% \small

%     \small
%     \begin{tabu}{llcccc|c}
%     \hline
%     \multirow{2}{*}{Model} & \multirow{2}{*}{Training Data} & R1 & R2 & R3 & AVQA & VQA v2\\
%     \cmidrule(lr){3-3} \cmidrule(lr){4-4} \cmidrule(lr){5-5} \cmidrule(lr){6-6} \cmidrule(lr){7-7}
%     & & val/test & val/test & val/test &  val/test & test-dev\\
%     \hline
%     \multirow{2}{*}{ClipBERT} & VQA v2 +VGQA & 21.39/20.45 & 19.29/20.06 & 21.51/22.63 & 20.69/20.82\\
%     %  & +R1 &  23.83/22.43 & 20.08/20.13 & & \\
%     %  & +R1+R2 &  24.03/23.08 & 23.12/23.86& & \\
%      & +AVQA & & & & \\
%     \hline
%     \hline
%     \multirow{2}{*}{VILLA-B} & VQA v2 +VGQA & 21.22/19.45 & 18.53/18.92 & 19.86/19.45 & \\
%     %  & +R1 & 25.92/24.07 & 20.00/20.05 & & \\
%     %  & +R1+R2 & 27.53/25.13 & 23.23/23.91 &  & \\
%      & +AVQA & 30.78/28.43 & 25.66/25.11 & 23.32/23.36 & 27.29/26.10\\
%     \hline
%     \multirow{2}{*}{VILLA-L} & VQA v2 +VGQA & 24.99/22.88 & 18.58/18.23  & 21.67/ \\
%     %  & +R1 & 28.29/26.12& 19.44/19.02& & \\
%     %  & +R1+R2 &30.02/27.81 & 24.05/23.59  & & \\
%      & +AVQA & & & & \\
%     \hline
%     \end{tabu}
% \vspace{-8pt}
%     \caption{\small{Evaluation of AT-base method VILLA~\cite{gan2020large} and grid feature based method ClipBERT~\cite{}. `AVQA' refers to R1+R2+R3.}}
%     \label{tab:at_feat}
%     \vspace{-3mm}
% \end{table*}

\begin{table}[t!]
% \tablestyle{1pt}{0.9}
    \centering
\small
\resizebox{0.8\linewidth}{!}{
    \begin{tabu}{ll|cc}
    \hline
    \multirow{2}{*}{Model} & \multirow{2}{*}{Training Data} & AVQA & VQA v2\\
    \cmidrule(lr){3-3} \cmidrule(lr){4-4} 
    & & test  & test-dev\\
    \hline
    \multirow{2}{*}{UNITER-B} & VQA v2 +VGQA & 18.81  & 72.70\\
     & ALL & 24.10 & 72.66\\
    % \hline
    \hline
    \multirow{2}{*}{ClipBERT} & VQA v2 +VGQA & 21.16 & 69.08\\
     & ALL & 24.35 & 69.17\\
    % \hline
    \hline
    \multirow{2}{*}{VILLA-B} & VQA v2 +VGQA & 19.68 & 73.37\\
     & ALL & \textbf{26.08} &\textbf{74.28}\\
    \hline
    \end{tabu}
}
\vspace{-8pt}
    \caption{\small{Evaluation of grid-feature-based method ClipBERT~\cite{lei2021less}, and adversarial-training-based method VILLA~\cite{gan2020large}. `ALL' refers to VQA v2+VGQA+AVQA.}}
    \label{tab:at_feat}
    \vspace{-6pt}
\end{table}
\begin{table*}[hbt!]
    \centering
    \small
% \resizebox{.99\textwidth}{!}{
    \begin{tabu}{llcccccc}
    \hline
    \multirow{2}{*}{Model} & \multirow{2}{*}{Training Data} & VQA-Rep. & \specialcell{VQA-LOL\\Comp.} & \specialcell{VQA-LOL\\Supp.} & VQA-Intro. & CV-VQA  & IV-VQA \\
    \cmidrule(lr){3-3} \cmidrule(lr){4-4} \cmidrule(lr){5-5} \cmidrule(lr){6-6} \cmidrule(lr){7-7} \cmidrule(lr){8-8}
    & & Acc. \textcolor{green}{$\uparrow$} %rephrase
& Acc. \textcolor{green}{$\uparrow$} %compose
& Acc. \textcolor{green}{$\uparrow$} %supplement
& M$\checkmark$ S$\checkmark$ \textcolor{green}{$\uparrow$} % intro
& \#flips \textcolor{red}{$\downarrow$} %IV-VQA
& \#flips \textcolor{red}{$\downarrow$}\\ %CV-VQA 
    \hline
    Previous models & VQA v2 Train & 56.59~\cite{shah2019vqa-rephrase} & 49.88~\cite{gokhale2020vqa-lol} & 50.54~\cite{gokhale2020vqa-lol} & 50.05~\cite{selvaraju2020vqa-introspect} & 7.53~\cite{agarwal2020causal-vqa} & 78.44~\cite{agarwal2020causal-vqa}\\
    \hline
    \multirow{1}{*}{UNITER-B}~\cite{li2020closer} & VQA v2 Train & 64.66 & 54.16 & 49.89 & 56.69 & 8.47 & 40.67\\
    \hline
    \multirow{2}{*}{UNITER-B (ours)} & VQA v2 Train & 64.56 & 54.54 & 50.00 & 56.80 & 8.44 & 39.97\\
     & +AVQA & \textbf{65.42} & \textbf{55.10} & \textbf{51.36} & \textbf{57.93} &  \textbf{8.43}& \textbf{38.40}\\
    \hline
    \end{tabu}
% }
\vspace{-8pt}
    \caption{\small{Model performance on recent robust VQA benchmarks.}}
    \label{tab:robust_eval}
    \vspace{-2mm}
\end{table*}
\begin{table*}[h!]
    \centering
\small
\resizebox{0.9\textwidth}{!}{
    \small
    \begin{tabu}{lccccccccccc}
    \hline
    \multirow{2}{*}{Round} & \multirow{2}{*}{Count} & \multirow{2}{*}{OCR} & \multicolumn{4}{c}{Reasoning} & \multicolumn{5}{c}{Visual Concept Recognition}\\
    \cmidrule(lr){4-7} \cmidrule(lr){8-12}
    & & & Position & Relation & \specialcell{Common-\\sense} & \specialcell{Other} & \specialcell{Low-\\level} & Action & \specialcell{Small\\Object} & Occlusion & Abstract \\
    \hline
    % R1 & 117 & 53 & 73 & 42 & 77 & 4 & 49 & 21 & 66 & 73 & 34 & 10\\ 
    R1 & 23.3\% & 10.7\% & 14.7\% & 8.3\% & 17.3\% & 0.7\% & 9.7\% & 4.3\% & 13.3\% & 14.7\% & 6.3\% \\ 
    % R2 & 45+45 & 32+36  & 26+10 & 49+34 & 41+19 & 9+4  & 18+20 & 21+7 & 24+34 & 16+14 &14+32  \\ 
    R2 & 30.0\% & 22.7\%  & 12.0\% & 27.7\% & 20.0\% & 4.3\% & 12.7\% & 9.3\% & 22.7\% & 10.0\% & 15.3\% \\
     % R3 & 106 & 39 & 39 & 85 & 75 & 19 & 35 & 13 & 60 & 60 & 18\\ 
    R3 & 35.3\% & 13.0\% & 13.0\% & 28.3\% & 25.0\% & 6.3\% & 11.7\% & 4.3\% & 20.0\% & 20.0\% & 6.0\%\\
    \hline
    % Ave. & 27.4\% & 14.7\% & 10.5\% & 14.8\% & 17.3\% & 3.7\% & 9.4\% & 4.3\% & 12.9\% & 12.5\% & 6.6\%\\
    Ave. & 29.6\% & 15.4\% & 13.2\% & 21.4\% & 20.8\% & 3.8\% & 11.3\% & 6.0\% & 18.7\% & 14.9\% & 9.2\%\\
    \hline
    \end{tabu}
}
\vspace{-8pt}
    \caption{\small{Analysis of 300 randomly sampled AVQA examples per round and on average. Low-level visual concepts include color, shape, and texture. A question may belong to multiple different categories.}}
    \label{tab:sample_analysis}
    \vspace{-6pt}
\end{table*}
We further investigate if the low performance is due to the difference in answer distribution between training and testing split. Due to the large number of answer candidates (more than 3000 for VQA v2), it is impossible to evenly balance the possibility of each answer. Therefore, we test out this hypothesis by adopting a simple yet effective baseline method on VQA-CP~\cite{teney2020value}: adding a regularization term by replacing the image with a randomly sampled one. The intuition is that the answer to a question corresponding to a given image is very unlikely to be correct for a randomly sampled image. As reported in Table~\ref{tab:vqa-cp}, although effective on VQA-CP, adding such regularization hurts the performance on AVQA for both BUTD and UNITER-B. In addition, when applied to a stronger model on VQA-CP, \emph{i.e.}, UNITER-B, the regularization term is less effective.

\vspace{2pt}
\noindent\textbf{\emph{Q2: Is AVQA transferrable to different visual features?}}
The AVQA dataset is collected with the assistance of models trained on Faster R-CNN~\cite{ren2015faster} region features~\cite{anderson2018bottom}.
To investigate whether these collected adversarial examples are transferrable to different image features, we conduct experiments using another type of feature, \emph{i.e.}, grid features~\cite{jiang2020defense} from CNNs, which have shown to be effective for VQA tasks~\cite{jiang2020defense,huang2020pixel,nguyen2020revisiting,lei2021less}. 
Specifically, we consider ClipBERT~\cite{lei2021less}, 
an end-to-end pre-trained model that directly takes in raw images and questions, and the images are represented by grid features as in~\cite{jiang2020defense}.
Meanwhile, ClipBERT's end-to-end training strategy may also help to defend potential attacks to fixed feature representations widely used in previous work~\cite{chen2019uniter,tan2019lxmert,anderson2018bottom}.
Table~\ref{tab:at_feat} compares ClipBERT against UNITER-B.  The poor performance of ClipBERT on AVQA suggests that adversarial examples in AVQA are transferrable to different image representations. However, ClipBERT performs comparably to UNITER-B on AVQA, although it significantly under-performs UNITER-B on VQA v2, which suggests that VQA v2 may not be  reliable for evaluating model robustness.

\begin{figure*}
     \centering
     \begin{subfigure}[b]{0.98\textwidth}
         \centering
         \includegraphics[width=0.98\textwidth]{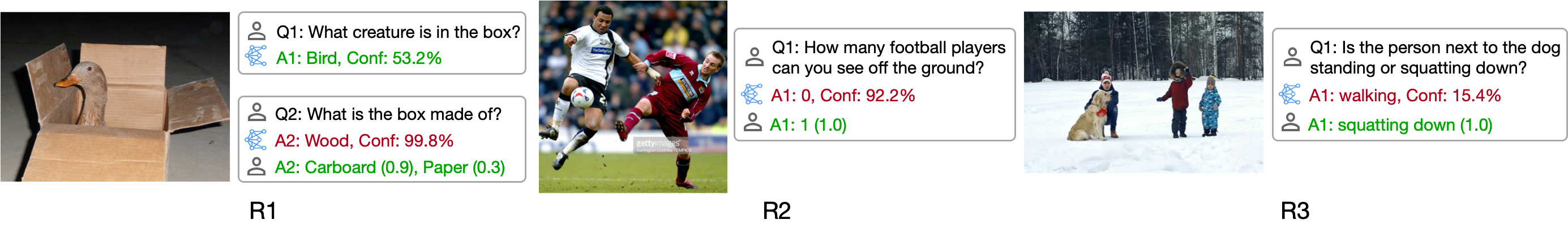}
         \caption{\small{Visualization of examples collected per round in AVQA. Each ground-truth answer (VQA score) is collected from 10 workers.}}
         \label{fig:vis_round}
     \end{subfigure}
     \hfill
     \vspace{1.5mm}
     \begin{subfigure}[b]{0.98\textwidth}
         \centering
         \includegraphics[width=0.98\textwidth]{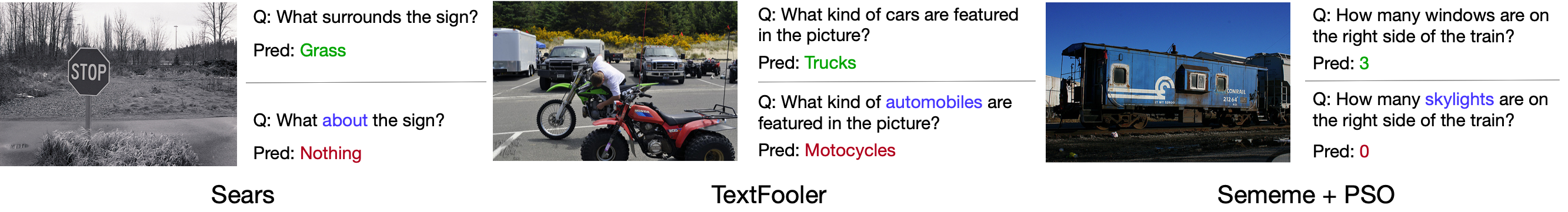}
         \caption{\small{Visualization of examples generated via textual adversarial attack methods. \textcolor{blue}{Blue} indicates the changes made in adversarial questions.}}
         \label{fig:adv_attack}
     \end{subfigure}
    \vspace{-2mm}
    \caption{\small {Illustration of adversarial examples from (a) AVQA and (b) textual adversarial attack methods: Sears~\cite{ribeiro-etal-2018-semantically}, Textfooler~\cite{jin2020bert} and Sememe+PSO~\cite{zang2020word}. \textcolor{green}{Green} (\textcolor{red}{red}) indicates a correct (wrong) answer.} }
    \label{fig:vis}
    \vspace{-8pt}
\end{figure*}

\vspace{2pt}
\noindent\textbf{\emph{Q3: How effective is adversarial training on AVQA?}} We examine the effectiveness of adversarial training by adopting PGD-based adversarial training method VILLA in ~\cite{gan2020large}. VILLA-B is both adversarially pre-trained on large-scale image-text data and adversarially finetuned on the respective dataset. We compare its performance against UNITER-B on both AVQA and VQA v2 in Table~\ref{tab:at_feat}. Adversarial training brings slight performance improvement. However, the performance gap between AVQA and VQA v2 is still very significant. Note that VILLA-B crafts adversarial examples during training by adding adversarial perturbations to the embedding space. These adversarial perturbations can hardly change the intrinsic statistics of training data, such as the distribution of question types and relevant objects in the image. Our analysis of question types and visual recognition concepts in Sec.~\ref{sec:q_type} will show that AVQA is hard because it requires the model to have the ability to reason, count and recognize different visual concepts. 

\subsection{Evaluation on Other Datasets}\label{sec:robust_eval}
We also test models on recent robust VQA benchmarks including: VQA-Rephrasings~\cite{shah2019vqa-rephrase} for linguistic variations, VQA-LOL~\cite{gokhale2020vqa-lol} Complement/Supplement for logical reasoning, VQA-Introspect~\cite{selvaraju2020vqa-introspect} for consistency of model predictions in perceptual sub-questions and main reasoning questions, CV-VQA~\cite{agarwal2020causal-vqa} and IV-VQA~\cite{agarwal2020causal-vqa} for model robustness to image manipulations. Results are summarized in Table~\ref{tab:robust_eval}. We observe that UNITER-B can already outperform previous models for most of the benchmarks, which is consistent with observations in~\cite{li2020closer}. Training on AVQA is helpful in improving model performance on robustness benchmarks.
Particularly, AVQA helps to boost model reasoning capability across 3 datasets. It is likely that AVQA exposes the model training to more diverse question templates, hence improves on VQA-Rephrasings. 
On IV-VQA, which focuses on counting questions, AVQA helps to improve performance despite of the significant performance gain UNITER-B has already achieved.  

\subsection{Analysis on Question Types}\label{sec:q_type}
We manually annotate 300 randomly sampled examples from each round to investigate: \textit{which types of questions do workers employ to fool the models}, and \textit{how they evolve as the rounds progress}.

Results are summarized in Table~\ref{tab:sample_analysis}. Questions are categorized into 4 meta-categories: counting, OCR, reasoning, and visual concept recognition. Although OCR and counting can be considered as visual concept and quantitative reasoning, we separate them out as they contribute a large portion per round, to almost 50\% in the later rounds. There are three main reasoning questions: positional reasoning (\emph{i.e.}, the relative/absolute position of an object), relational reasoning (\emph{i.e.}, semantic relationship between two or more objects), and commonsense reasoning (\emph{i.e.}, visual commonsense reasoning, \emph{e.g.}, ``Is the water more likely to be a lake or an ocean", given an image showing a body of water surrounded by mountains.). Other reasoning questions include comparative reasoning (\emph{e.g.}, ``which person is taller?") and logical reasoning (\emph{e.g.}, negation). For visual concept recognition, we roughly divide them into low-level visual concepts (\emph{e.g.}, color, shape, texture), action (\emph{e.g.}, ``what is the person doing"), small objects, occluded objects, and abstract objects (\emph{e.g.}, objects in painting). 

We observe that annotators rely heavily on counting questions to attack the models -- nearly 30\% of the sampled questions across all rounds fall into this category. While R1 questions are mostly on objects that are of normal sizes and less occluded, we found that the counting questions become harder in R2 and R3 as many of them are about small and occluded objects. There is also a surge in abstract and OCR questions for R2, due to the increase in the number of abstract images and images that contain scene text. The percentage of reasoning questions, especially relational reasoning and commonsense reasoning, increases drastically from R1 to R2 and R3. Visualizations in Figure~\ref{fig:vis_round} show that questions in later rounds are indeed more complicated, with more detailed relational and positional descriptions when referring to an object. 
Overall, these findings are compatible with the idea that VQA models are not robust enough to various types of questions.

\begin{table}[t!]
    \centering
\small
\resizebox{\linewidth}{!}{
\begin{tabu}{l|cccc}
    \hline
    Method  & \#Tries &  Error Rate & Orig. Acc. & Adv. Acc. \\
    \hline
    Sears~\cite{ribeiro-etal-2018-semantically} & 3.0 & 11.6\% & 69.1 & 63.0 \\
    Textfooler~\cite{jin2020bert} & 39.5 & 1.4\% & 69.1 &  67.8\\
    Sememe+PSO~\cite{zang2020word}$^\dagger$ & 35.9 & 88.6\% & 84.9 & 12.5 \\
    \hline
    AVQA & 1.6 & 38.1\% & - & - \\
    \hline
\end{tabu}
}
\vspace{-8pt}
    \caption{\small{Comparison to adversarial attack methods. Orig. Acc. (Adv. Acc.) is the accuracy on original (adversarial) examples. ($\dagger$) Note that Sememe+PSO only attacks questions longer than 10 words, so 94.8\% examples are not being attacked. }}
    \label{tab:adv_attack}
    \vspace{-6pt}
\end{table}

\subsection{Why Human-in-the-Loop?}\label{sec:attack_methods}
Textual adversarial attack methods~\cite{morris2020textattack,jin2020bert,zang2020word} have been widely explored in NLP. The goal is to alter model predictions with minor changes to the input textual queries, so that adversarial examples can be generated and model vulnerabilities can be identified automatically. We investigate whether we can directly apply these methods to generate adversarial examples in high quality and compare the generated examples to AVQA.  In total, we consider 3 different textual adversarial attack methods, including 
Sears~\cite{ribeiro-etal-2018-semantically} via bask-translation for sentence-level attacks, Textfooler~\cite{jin2020bert} and Sememe+PSO~\cite{zang2020word} by replacing words with its synonyms or words that share the same sememe annotations for word-level attacks.  The adversarial attacks are performed to all questions on 5000 images in the Karpathy split~\cite{karpathy2015deep}.  We visualize examples in Figure~\ref{fig:adv_attack}. Without human-in-the-loop, the generated adversarial questions share similar problems: ($i$) the adversarial question does not share the same answer with the original question, therefore additional answer annotations may need to be collected; ($ii$) model prediction to the adversarial question is not necessarily incorrect when it is different from answers to the original question; ($iii$) word similarity may not hold when it needs to be grounded to the image (\emph{e.g.}, window vs. skylights).  In addition, we compare these methods against the AVQA dataset quantitatively in Table~\ref{tab:adv_attack}. Generally, humans take much fewer tries and have a higher successful rate when attacking VQA models. How to design effective adversarial attack methods to generate high-quality VQA examples can be an interesting future research direction. 
%------------------------------------------------------------------------
\vspace{-2pt}
\section{Conclusion}
\vspace{-1mm}
In this work, we collect a new benchmark Adversarial VQA (AVQA) to evaluate the robustness of VQA models.  
It is collected iteratively for 3 rounds via a human-and-model-in-the-loop enabled training paradigm, on images from different domains. AVQA questions cover diverse robustness types, enabling a more comprehensive evaluation on model robustness. 
Our analysis shows that state-of-the-art models cannot maintain decent performance on AVQA, despite of large-scale pre-training, adversarial training, sophisticated model architecture design, and stronger visual features.  AVQA brings a new challenge to the community on how to design more robust VQA models that are ready to deploy in real-life applications.

{\small
\bibliographystyle{ieee_fullname}
\bibliography{egbib}
}

% for arxiv purposes; commented out for main text only
\clearpage
\appendix
\section{More Discussions on AVQA}
\paragraph{Future Practices.} We recommend future models to report performance on both VQA v2 and AVQA. AVQA is designed to test VQA model robustness under human adversarial attacks. It is complementary to VQA v2 (naturally-collected questions), rather than a replacement. In addition, we believe it is beneficial to evaluate on other robust VQA benchmarks as well.
While AVQA encompasses broader robustness types and image domains with higher data quality, existing robustness benchmarks can in addition provide useful analysis tailored to individual robustness types.  An ideal VQA system should perform well on all VQA benchmarks. 
Further, we encourage future work to apply human-in-the-loop adversarial attack to their proposed models to identify potential vulnerabilities. For AVQA, we expect to provide a dynamically evolving VQA benchmark as models grow more robust, to alleviate the drawbacks of static benchmarks (\emph{e.g.}, performance saturation and overfitting).

\paragraph{Constraints/Rules For Data Collection.} As our goal is to examine VQA models' robustness when encountering test examples in the wild, we do not constrain the questions to specific types, to avoid unconscious bias from dataset creators. 
As a result, we have found that models make more mistakes on Count/OCR/Relation/Commonsense questions (Table 7 in the main text).

To obtain high-quality adversarial questions, we enforce a set of rules to ensure the questions are objective, relevant to the image, and have exact answers (see detailed instructions in Figure~\ref{fig:q_instructions}). We also manually filter out questions with repetitive patterns for each annotator during collection. Our answer annotation process validates the collected questions to some extent, which are answerable by human but not always answerable by model.

\paragraph{Bias in Model Choices.} Our current choice of models was guided by the assumption that newer models are more likely to be transformer-based with currently-proven most effective features. We plan to include a broader choice of models in future collection, as the benchmark evolves.

\begin{table*}[h]
    \centering
\small

    \small
    \begin{tabu}{lcccc|cc|c}
    \hline
    \multirow{2}{*}{Round} & \multicolumn{4}{c}{Question Types} & \multicolumn{2}{c}{Upper Bound} & Human Performance\\
    \cmidrule(lr){2-5} \cmidrule(lr){6-7} \cmidrule(lr){8-8}
    & Y/N & Num & OOV$^\star$ &  Other & val/test$^\star$ & val/test$^\dagger$ & val/test\\
    \hline
    R1 & 13.53\% & 23.36\% & 10.03\% & 50.08\% & 81.43/79.75 & 92.03/92.05 & 74.92/75.14  \\
    R2 & 8.62\% & 29.91\% & 14.37\% & 47.01\% & 76.26/77.11 & 93.60/93.43 & 78.29/78.83 \\
    R3 & 11.24\% & 35.55\% & 12.17\% & 41.04\% & 79.64/80.91 & 94.48/94.41  & 81.61/81.15\\
    \hline
    % AVQA & 11.51\% & 27.89\% & 11.90\% & 48.71\% & 79.32/79.11 &  93.19/ 93.02\\
    AVQA & 11.40\% & 28.90\% & 11.95\% & 47.75\% & 79.27/79.28 & 93.28/93.21 & 78.05/78.15\\
    \hline
    \end{tabu}
\vspace{-8pt}
    \caption{\small{Question type distribution on verified examples and upper bound on val/test set across three rounds. $\star$ is based on VQA v2~\cite{goyal2017making} answer vocabularies. $\dagger$ is based on open vocabularies.}}
    \label{tab:q_type}
    \vspace{-3mm}
\end{table*}

\section{Data Statistics}
\begin{figure}[t!]
         \centering
         \includegraphics[width=\linewidth]{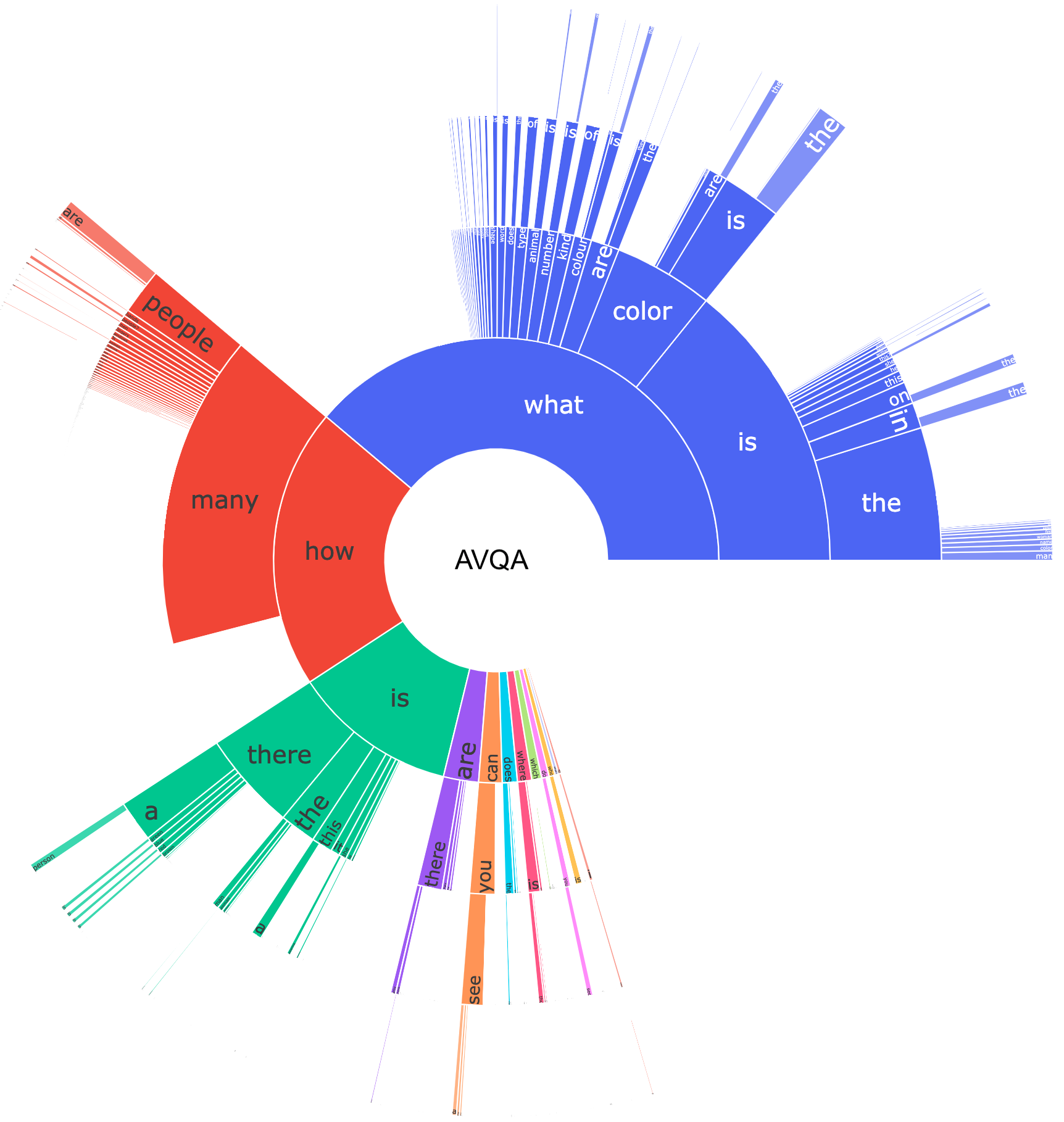}
         \caption{Distribution of questions by their first four words. The arc length is proportional to the number of questions containing the word. White areas are words with contributions too small to show.}
         \label{fig:q_dist}
         \vspace{-3mm}
\end{figure}

\paragraph{Type of Questions.} Following~\cite{antol2015vqa}, given the structure of questions generated in English, we cluster questions into different types based on the words that start the question. Figure~\ref{fig:q_dist} shows the distribution of questions based on the first four words of the questions in AVQA. Interestingly, the variety of question types are quite similar to those in~\cite{antol2015vqa}, including “What is”, “How many” and “Is there”. Quantitatively, we also categorize the questions into ``Y/N", ``Num", ``OOV" and ``Other". The percentage of questions for different categories
is shown in Table~\ref{tab:q_type}. ``OOV" questions refer to questions that cannot be answered by VQA v2~\cite{goyal2017making} answer vocabularies. We also include two upper bounds, one based on VQA v2 answer vocabularies, and the other on open vocabularies. Moreover, we estimate human performance on AVQA by sampling 1 human answer as prediction and use the rest 9 answers as references. We repeat the process 10 times and average the score. Comparing model performance reported in the main text, there is still a huge gap, with about 50 points lower than the upper bounds or the estimated human performance. 

\begin{figure}[t!]
         \centering
         \includegraphics[width=\linewidth]{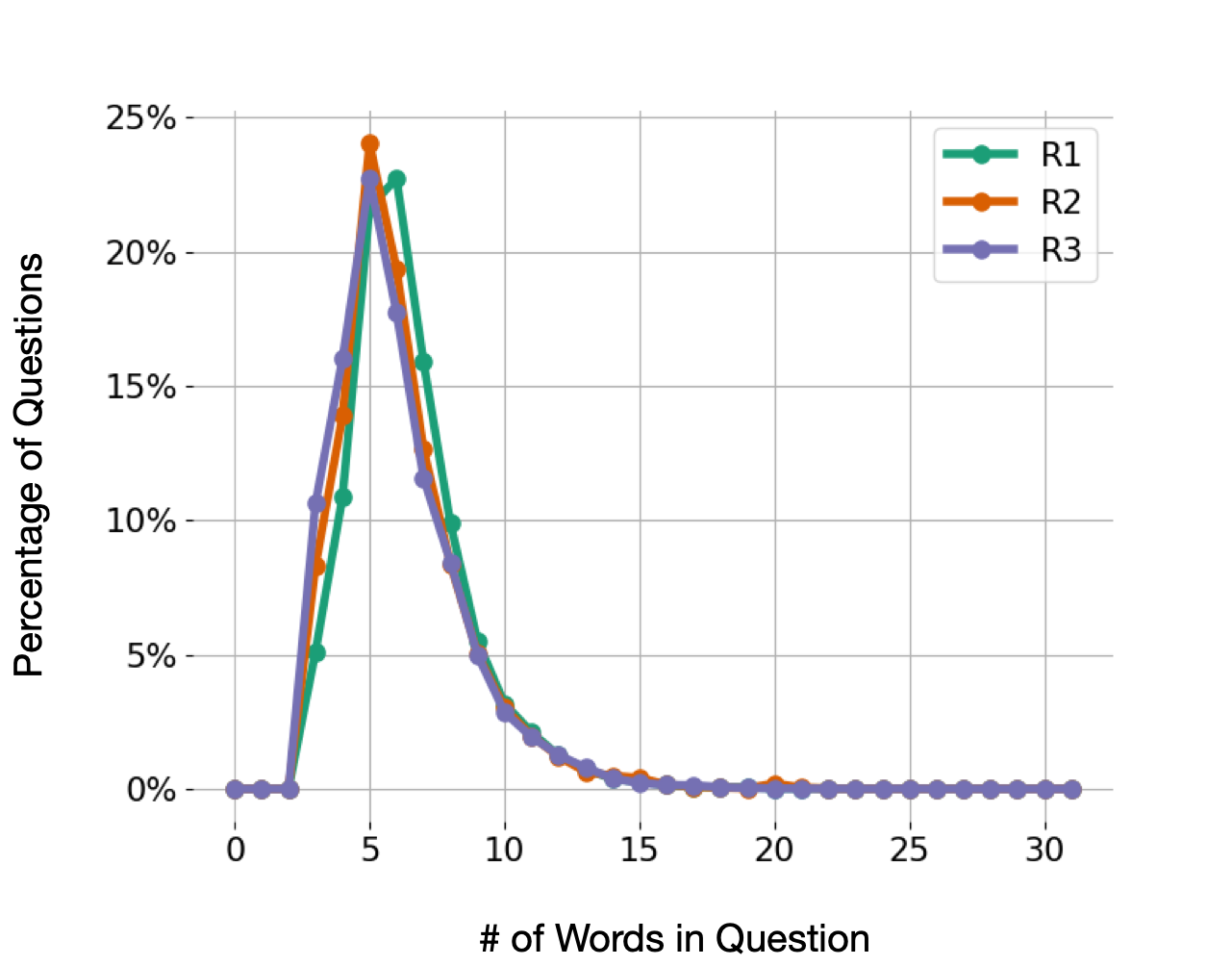}
         \caption{Percentage of questions with different word length across three rounds. Most questions range from four to ten words.}
         \label{fig:q_len}
         \vspace{-3mm}
     \end{figure}

\paragraph{Question Lengths.} Figure~\ref{fig:q_len} shows the distribution of question lengths. We see that most questions range from four to ten words.

\begin{figure*}
    \centering
    \includegraphics[height=1.8in]{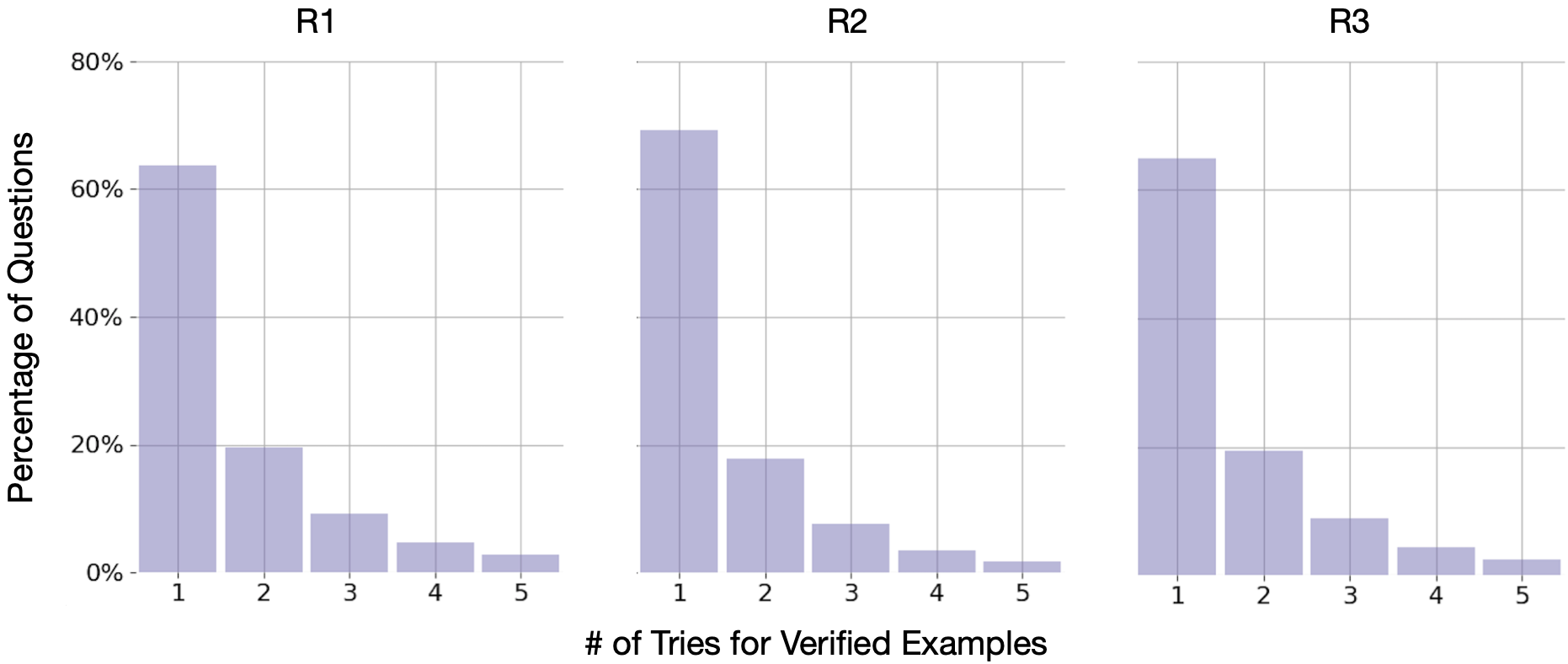}
    \caption{Histogram of the number of tries for each good verified example across three rounds.}
    \label{fig:num_tries}
\end{figure*}

\begin{figure*}
    \centering
    \includegraphics[height=2in]{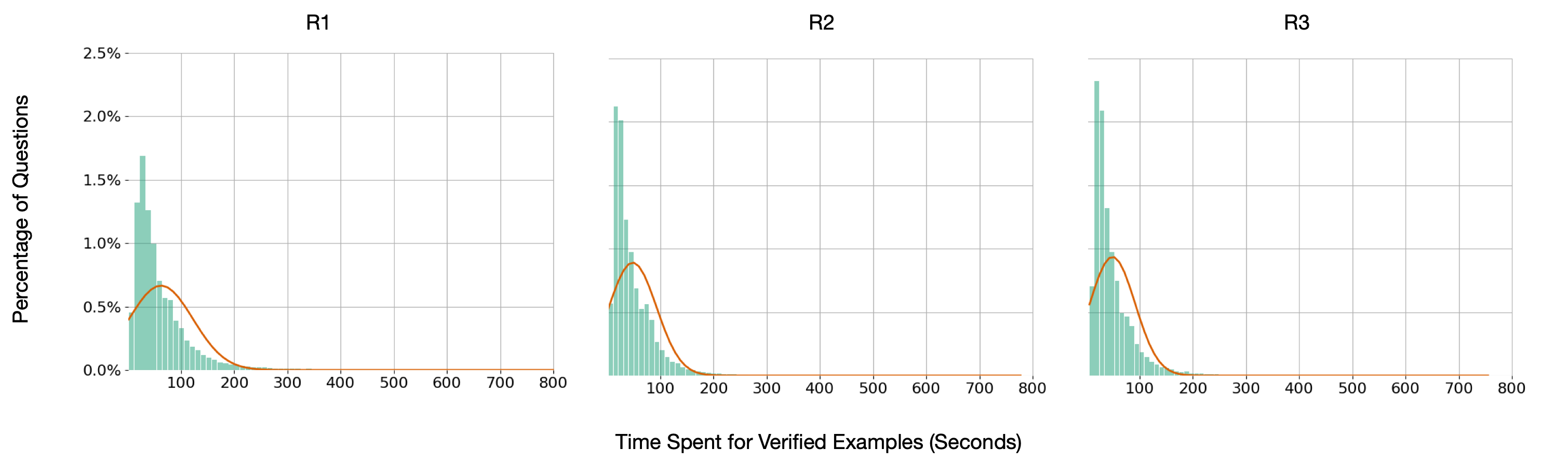}
    \caption{Histogram of the time spent per good verified example across three rounds.}
    \label{fig:timecost}
\end{figure*}

\begin{figure*}
    \centering
    \includegraphics[width=\textwidth]{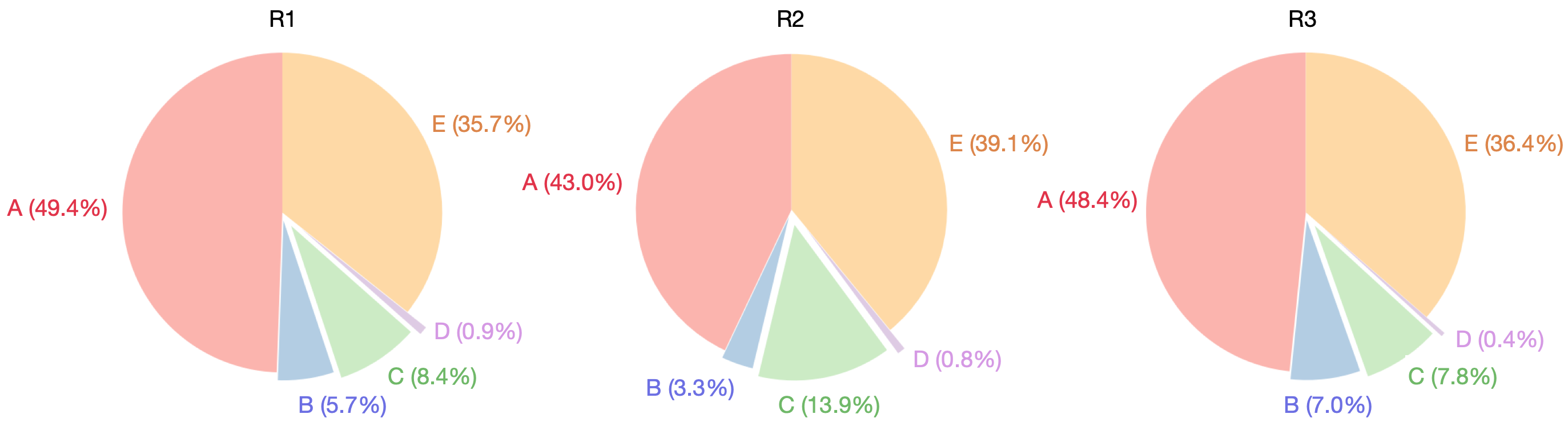}
    \caption{Proportion across three rounds. \textbf{A}=Examples that model got right (``Definitely Correct") during question collection, \textbf{B}=Examples that model neither got right nor wrong (``Not Sure") during question collection. \textbf{C}, \textbf{D} and \textbf{E} are examples that model got wrong (``Definitely Wrong") during question collection and sent to 9 annotators for verification during answer collection. Specifically, \textbf{C}=Examples that more than 3 verifiers overruled the question author's decision of ``Definitely Wrong'' and agree with the model's answer.
    % , meaning that the model answer receives score 1
    \textbf{D}=Examples for which there is no agreement among verifiers, \textbf{E}=Examples where at least two verifiers agree with each other during answer collection. We split \textbf{E} by images into training, validation, and testing sets. Examples on training images in \textbf{A} and \textbf{C} are added to the training set, the rest are discarded.  \textbf{B} and \textbf{D} are excluded due to low quality.}
    \label{fig:proportion}
\end{figure*}

\paragraph{Dataset Properties Across Rounds.} Figure~\ref{fig:num_tries} shows
a histogram of the number of tries per verified example across the three different rounds. We observe a consistent trend for all three rounds, over 80\% of examples are successfully collected within 2 tries. Figure~\ref{fig:timecost} shows the time taken per verified example. As the round progresses, we observe that more and more examples are collected within 100 seconds (less than 2 minutes). Figure~\ref{fig:proportion} shows the proportion of different types of collected examples across three rounds. Comparing to R1 and R2, R3 contains more ``not sure" judgements to model answers during question collection (type \textbf{B}), which indicates that the task is getting harder. There are a small amount of examples in all three rounds that there is no agreement among the answers collected (type \textbf{D}). Examples from \textbf{B} an \textbf{D} are excluded due to low quality. The rest are split into train/val/test set (refer to Figure~\ref{fig:proportion} captions for more information).


\begin{figure*}
    \centering
    \includegraphics[width=\textwidth]{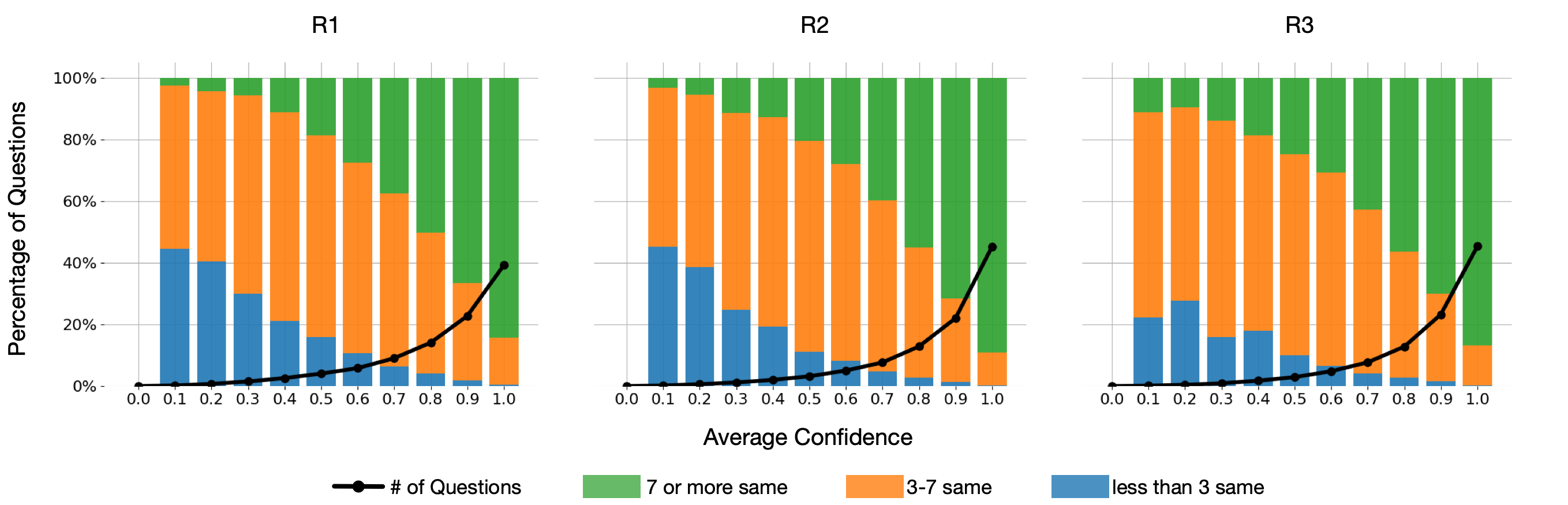}
         \caption{\small{ Number of questions per average confidence score across three rounds (black lines, 0 = not confident, 1 = confident). Percentage of questions where 7 or more answers are same, 3-7 are same, less than 3 are same across three rounds (color bars).}}
    \label{fig:ans_conf}
\end{figure*}

\paragraph{Answer Confidence and Inter-human Agreement.} During answer collection (see interface in Figure~\ref{fig:ans_interface}), the annotators are required to provide both a correct answer to the question given the image content and a self-judgment on how confident they feel about the answer. Specifically, we ask ``Do you think you were able to answer the question correctly?", and the annotator need to choose from ``yes" (confident with score 1), ``maybe" or ``no" (not confident with score 0). Figure~\ref{fig:ans_conf} shows the distribution of responses (black lines). A majority of the answers were labeled as confident. More than 9 annotators are confident about their answers on over 60\% questions on average.

In addition, we investigate how the self-judgment confidence corresponds to the answer agreement between annotators across three rounds of data collection. Color bars in Figure~\ref{fig:ans_conf} show the percentage of questions in which ($i$) 7 or more, ($ii$) 3-7, or ($iii$) less than 3 workers agree on the answers given their average confidence score. Across all rounds, the agreement between subjects increases with confidence.   We do observe that workers are more confident about their answers in R2 and R3, comparing to R1.

\begin{figure*}
    \centering
    \includegraphics[width=\textwidth]{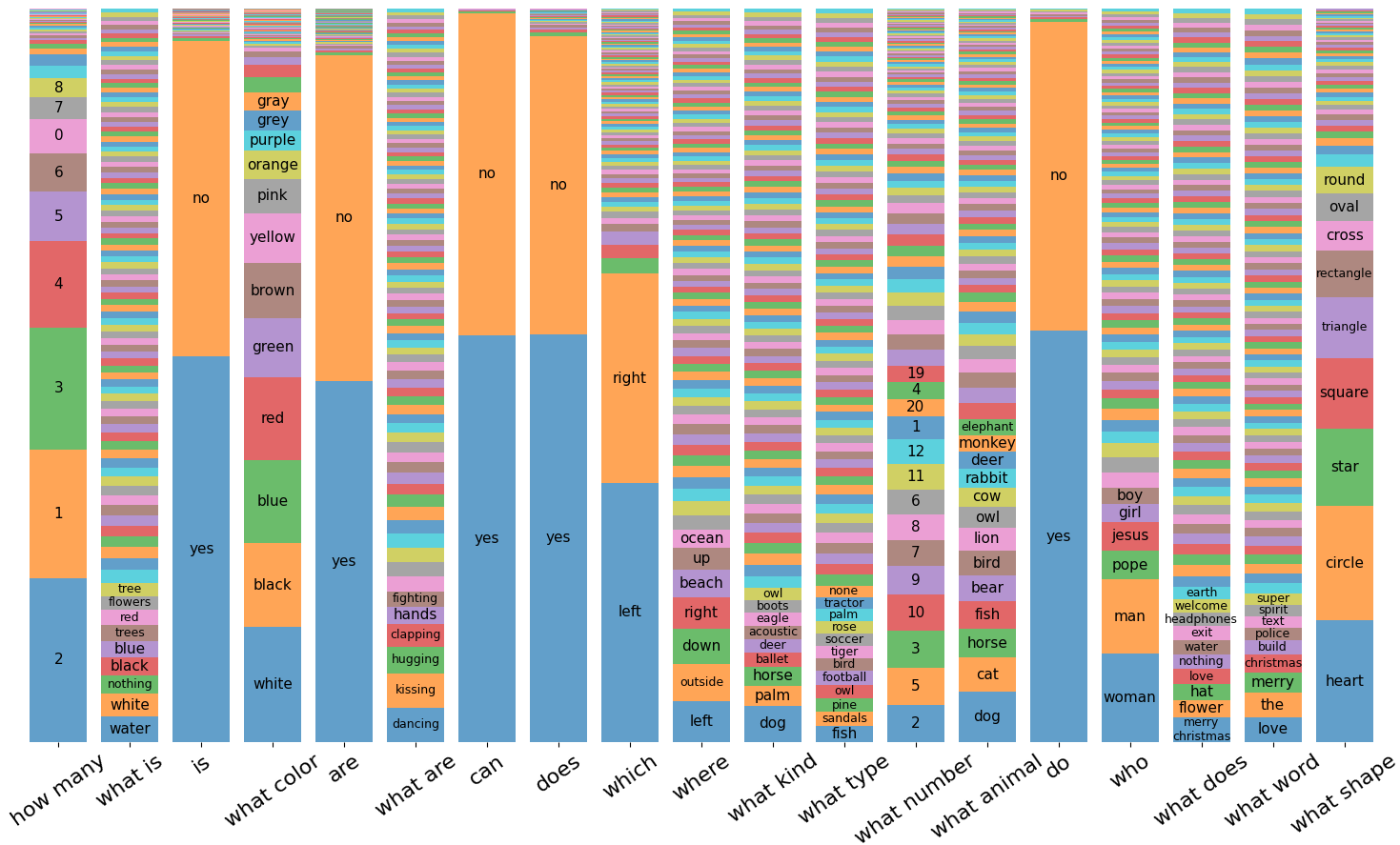}
    \caption{Distribution of answers per question type. Only top-100 answers to each question type are plotted. The height of each color bar is proportional to the percentage of an answer to the corresponding question type.}
    \label{fig:ans_dist}
\end{figure*}

\paragraph{Answer Distribution.} Figure~\ref{fig:ans_dist} shows the distribution of answers for several question types. We can see that a number of question types, such as “Is . . . ”, “Can. . . ”, and “Does. . . ” are typically answered using “yes” and “no” as answers. Other
questions such as “What is/are. . . ” and “What kind/type. . . ” have a rich diversity of responses. Other question types such as “What color. . . ” or “Which. . . ” have more specialized responses, such as colors, or “left” and “right”. These observations are similar to those in VQA v2.

\section{More Visualizations}
\begin{figure*}
    \centering
    \includegraphics[width=\textwidth]{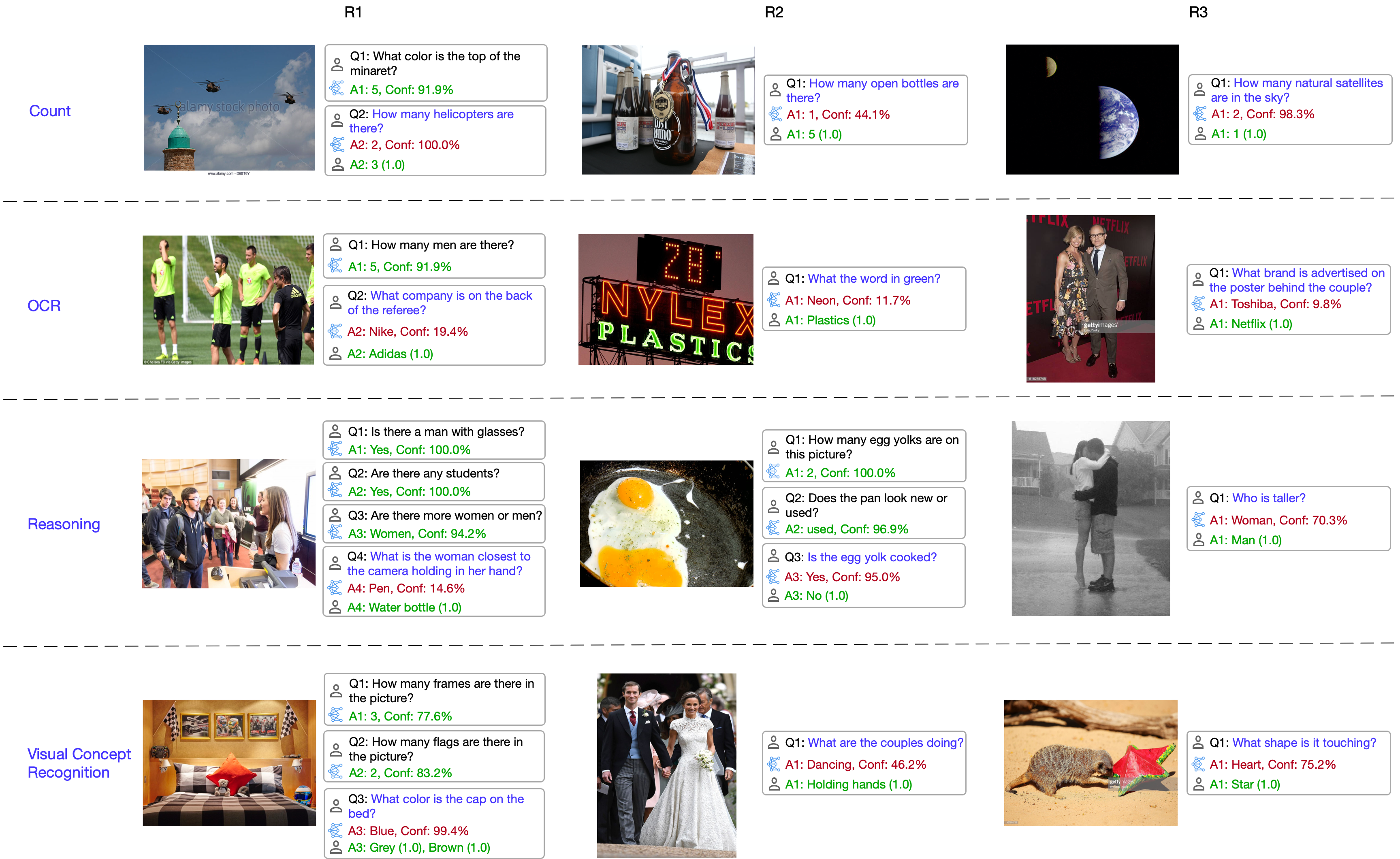}
    \caption{More visualization of examples collected per round in AVQA. We show examples that contains adversarial questions from 4 categories: Count, OCR, Reasoning and Visual Concept Recognition across three rounds. Each ground-truth answer (VQA score) is collected from 10 workers. \textcolor{green}{Green} (\textcolor{red}{red}) indicates a correct (wrong) answer. \textcolor{blue}{Blue} highlights the verified adversarial questions.}
    \label{fig:round_vis}
\end{figure*}

We include more visualization examples of collected data across three rounds in Figure~\ref{fig:round_vis}. We show adversarial questions from 4 categories: Count, OCR, Reasoning and Visual Concept Recognition. Note that questions may belong to multiple categories. For example, counting question from R3 (``How many natural satellites are in the sky?") requires commonsense about ``natural satellites". OCR question from R1 (``What company is on the back of the referee?") not only requires commonsense about ``referee" but relational reasoning about ``on the back of". Reasoning questions include positional/relational reasoning (\emph{e.g.}, ``What is the woman closest to the camera holding in her hand?"), commonsense reasoning (\emph{e.g.}, ``Is the egg yolk cooked?") and comparative reasoning (``Who is taller?"). There are also questions that require recognition of both low-level visual concepts (\emph{e.g.}, color/shape) and high-level visual concepts (\emph{e.g.}, action, relation).

We also visualize more examples generated via textual adversarial attack methods (Sears~\cite{ribeiro-etal-2018-semantically}, Textfooler~\cite{jin2020bert} and Sememe+PSO~\cite{zang2020word}) in Figure~\ref{fig:adv_attack}. The first two columns show invalid examples, and the last column includes valid examples, based on our manual examination. Recall that our goal is to collect high-quality adversarial questions that can be used to \emph{accurately, thoroughly evaluate and examine} the weakness of VQA models. 
Automatically generated adversarial questions are often incorrect (requiring additional human efforts to validate their correctness), and limited to linguistic variations to existing questions, thereby they are unlikely to provide a comprehensive analysis.
\begin{figure*}
    \centering
    \includegraphics[width=\textwidth]{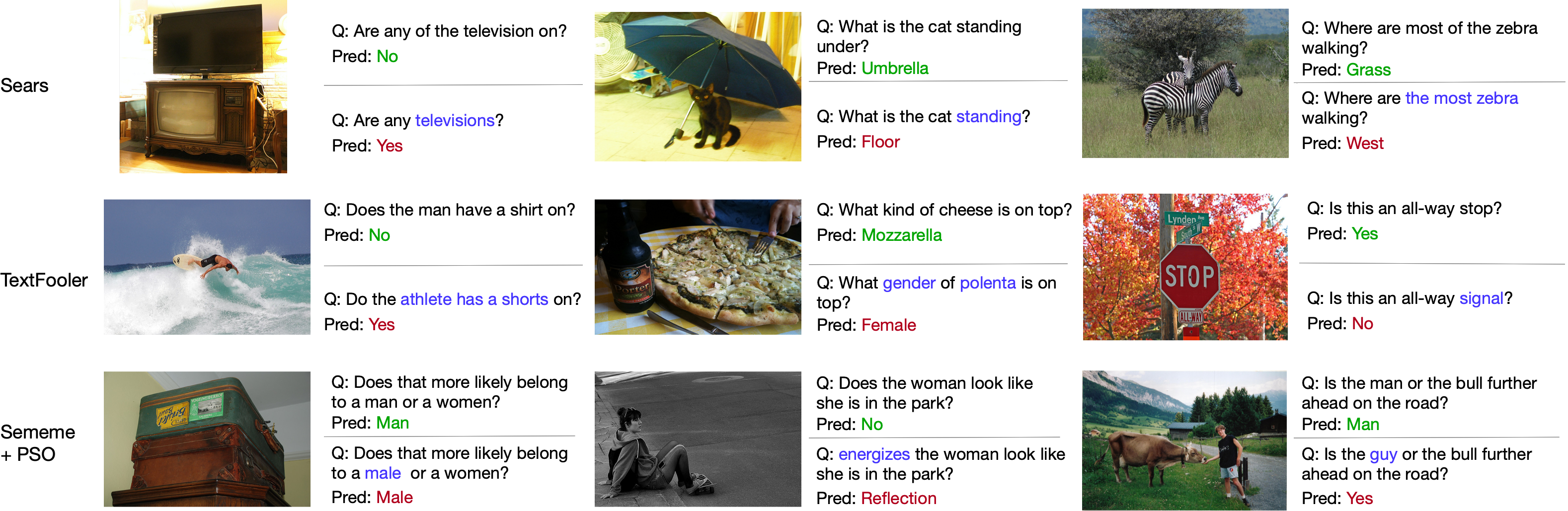}
    \caption{More adversarial examples from textual adversarial attack methods: Sears~\cite{ribeiro-etal-2018-semantically}, Textfooler~\cite{jin2020bert} and Sememe+PSO~\cite{zang2020word}. \textcolor{green}{Green} (\textcolor{red}{red}) indicates a correct (wrong) answer. \textcolor{blue}{Blue} highlights the changes made in adversarial questions.}
    \label{fig:adv_attack}
\end{figure*}

\section{More Results}
Recall that questions in R3 are collected on images from various domains, including web images from Conceptual Captions~\cite{sharma2018conceptual} (CC, used in R1 and R2), user-generated images from Fakeddit~\cite{fakeddit} and movie video frames from VCR~\cite{zellers2019vcr}. Hence, we can study how model performance can be transferable across different domains. We create a new split of R3 (R3$^\star$) according to the image source, with CC images for training and Fakeddit/VCR images for evaluation. Table~\ref{tab:domain_transfer} summarizes UNITER-B performance under different training settings. Despite the domain differences in images, the performance on Fakeddit and VCR split improves as we include more training data from CC images. Comparing the new split R3$^\star$ with the original split R3, training on more in-domain examples on CC images does help to improve model performance on R1 and R2. We also observe that model performance on VCR is significantly higher than those on the original R3 val and Fakeddit splits across all training settings. Images from VCR are often human-centric, which may be ``easier" than complex or abstract scenes depicted in CC/Fakeddit images.
\begin{table}[t!]
    \centering
\small
\resizebox{\linewidth}{!}{
    \small
    \begin{tabu}{lccc|cc}
    \hline
     Training Data & R1 & R2  & R3  &  Fakeddit~\cite{fakeddit} & VCR~\cite{zellers2019vcr}\\
    \hline
    VQA v2+VGQA & 20.60 & 17.86 & 20.71 & 19.59 & 23.34\\
    +R1 & 26.03 & 17.30 & 20.56 & 20.27 & 23.84\\
    +R2 & 26.60 & 23.21 & 19.26 & 17.85 & 22.05\\
    +R3$^\star$ & \textbf{27.02} & \textbf{23.78} & - & \textbf{22.56} & \textbf{27.43}\\
    \hline
    ALL & 26.85 & 23.38  & \textbf{24.48} & - & -\\
    \hline
    \end{tabu}
}
\vspace{-8pt}
    \caption{Domain transfer evaluation on UNITER-B. $\star$ indicates that we only use examples collected on CC~\cite{sharma2018conceptual} images for training. ALL refers to VQA v2+VGQA+R1+R2+R3.}
    \label{tab:domain_transfer}
    \vspace{-3mm}
\end{table}

In addition, we include detailed results from BUTD~\cite{anderson2018bottom}, ClipBERT~\cite{lei2021less}, VILLA-B and VILLA-L~\cite{gan2020large}  in Table~\ref{tab:more_results}. These results are consistent with observations we summarized in Section 4 of the main text.
\begin{table*}[t!]
    \centering
\small

    \small
    \begin{tabu}{llccccc|c}
    \hline
    \multirow{2}{*}{Model} & \multirow{2}{*}{Training Data} & R1 & R2 & R3 & AVQA & VQA v2 & $\Delta$(v2, AVQA)\\
    \cmidrule(lr){3-3} \cmidrule(lr){4-4} \cmidrule(lr){5-5} \cmidrule(lr){6-6} \cmidrule(lr){7-7} \cmidrule(lr){8-8}
    & & val/test & val/test & val/test &  val/test & test-dev & test-dev, test\\
    \hline
    \multirow{2}{*}{BUTD} & VQA v2 +VGQA & 20.80/19.28& 18.77/18.85 & 20.63/21.10 & 20.12/19.71 & 67.60 & 47.89 \\
     & +R1 & 20.27/20.27& 19.53/20.14 & 21.55/21.86 & 20.44/20.72 & 67.37 & 46.65\\
     & +R1+R2 & 24.41/21.82 & 22.28/21.80& 21.31/21.60 & 22.78/21.75 & 67.44 & 45.69\\
     & ALL & 24.96/22.11 & 22.62/22.78 & 23.92/23.61 & 23.91/22.78 & 67.52 & \textbf{44.74}\\
    \hline
     \hline
     \multirow{4}{*}{ClipBERT} & VQA v2 +VGQA & 21.39/20.45 & 19.29/20.06 & 21.01/23.16 & 20.45/21.16 & 69.08 & 47.92\\
     & +R1 &  23.83/22.43 & 20.08/20.13 & 22.49/22.65 & 22.25/21.78 & 69.07 & 47.29 \\
     & +R1+R2 &  24.03/23.08 & 23.12/23.86& 24.67/23.37 & 23.95/23.86 & 69.19 & 45.33\\
     & ALL & 24.62/23.68 & 22.96/24.66 & \textbf{25.05}/\textbf{24.87} & 24.24/24.35 & 69.17 & 44.82\\
    \hline
    \hline
    \multirow{4}{*}{VILLA-B} & VQA v2 +VGQA & 21.22/19.45 & 18.53/18.92 & 20.57/20.73 & 20.18/19.68 & 73.37  & 53.69\\
     & +R1 & 25.92/24.07 & 20.00/20.05 & 21.61/21.23 & 22.74/21.93 & 73.21 & 51.28\\
     & +R1+R2 & 27.53/25.13 & 23.23/23.91 & 21.96/21.87 & 24.46/23.74 & 73.11 & 49.37\\
     & ALL & \textbf{30.78}/\textbf{28.43} & \textbf{25.66}/\textbf{25.11} & 24.00/24.18 & \textbf{27.08}/\textbf{26.08} & 74.28 & 48.20\\
    \hline
    \multirow{4}{*}{VILLA-L} & VQA v2 +VGQA & 24.99/22.88 & 18.58/18.23  & 20.07/19.64 & 21.47/20.42 & \textbf{74.58} & 54.16\\
     & +R1 & 28.29/26.12& 19.44/19.02& 20.25/20.25 & 23.04/22.08 &  74.12 & 52.04\\
     & +R1+R2 &30.02/27.81 & 24.05/23.59  & 19.38/20.50 & 24.85/24.24 & 74.06 & 49.82\\
     & ALL & 29.92/28.01 & 24.59/24.26 & 23.66/23.09 & 26.32/25.32 & 74.24 & 48.92\\
    \hline
    \end{tabu}
\vspace{-5pt}
    \caption{Detailed results from BUTD~\cite{anderson2018bottom}, ClipBERT~\cite{lei2021less}, VILLA-B and VILLA-L~\cite{gan2020large} under different settings. AVQA = R1+R2+R3, ALL = VQA v2+VGQA+AVQA.}
    \label{tab:more_results}
    \vspace{-3mm}
\end{table*}

\section{Data Collection Interface}
\begin{figure*}
    \centering
    \includegraphics[width=.8\textwidth]{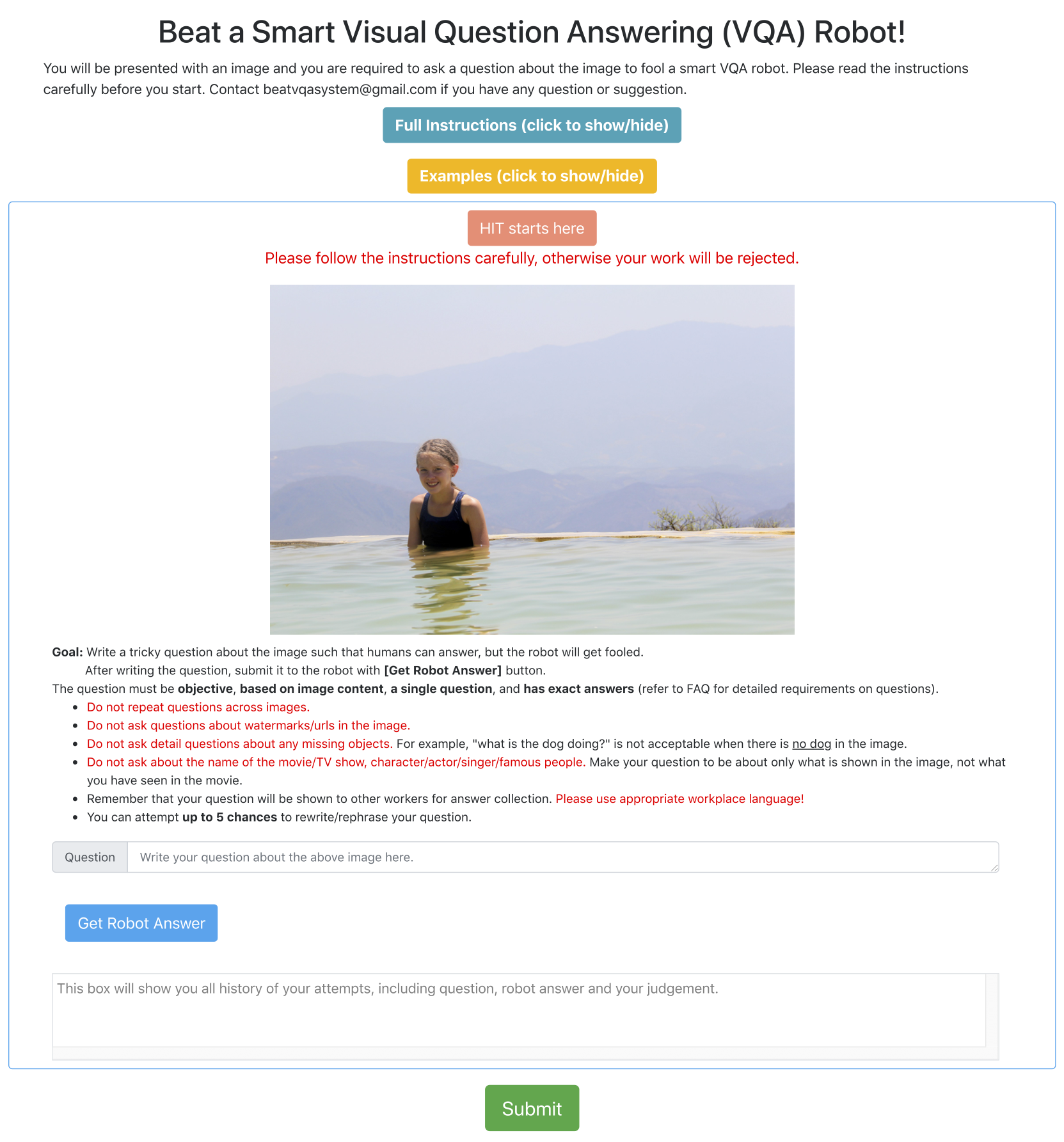}
    \caption{UI for question collection. Given an image, the annotator is required to write a tricky answer to fool our ``smart VQA robot'' (well-trained VQA models). After clicking the ``Get Robot Ansner'', the annotated question will be sent to our online model for evaluation, and a feedback will be returned immediately. See Figure~\ref{fig:q_interface_step2} for an example of model feedback.}
    \label{fig:q_interface_step1}
\end{figure*}

\begin{figure*}
    \centering
    \includegraphics[width=.8\textwidth]{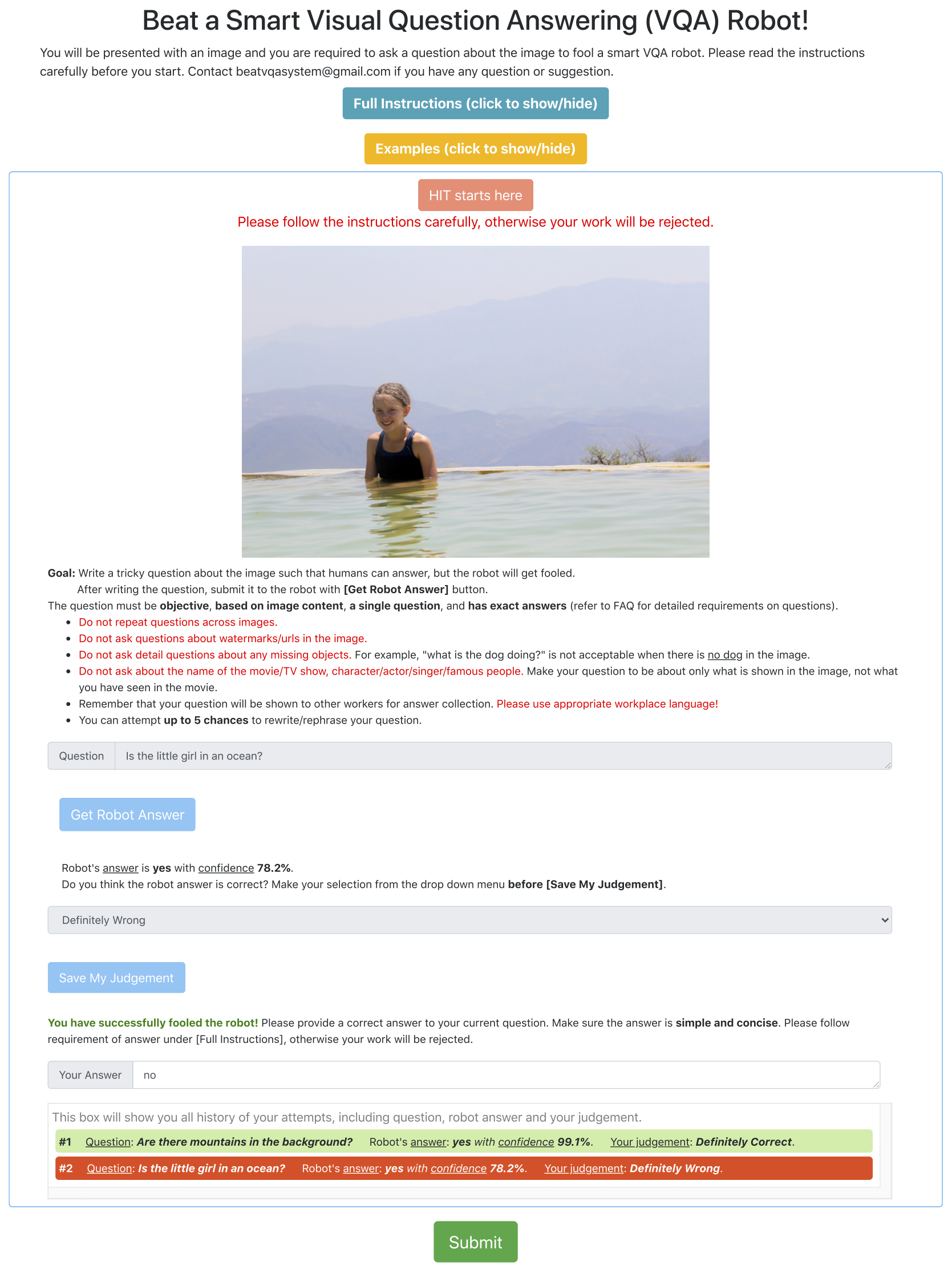}
    \caption{Example of model feedback shown to the annotators. After reviewing the model response, the annotator need to judge the correctness of the model answer (``Definitely  Correct", ``Not Sure", or ``Definitely Wrong"). If the model answer is definitely wrong, the annotator is prompted to enter a correct answer.}
    \label{fig:q_interface_step2}
\end{figure*}

\begin{figure*}
    \centering
    \includegraphics[width=.8\textwidth]{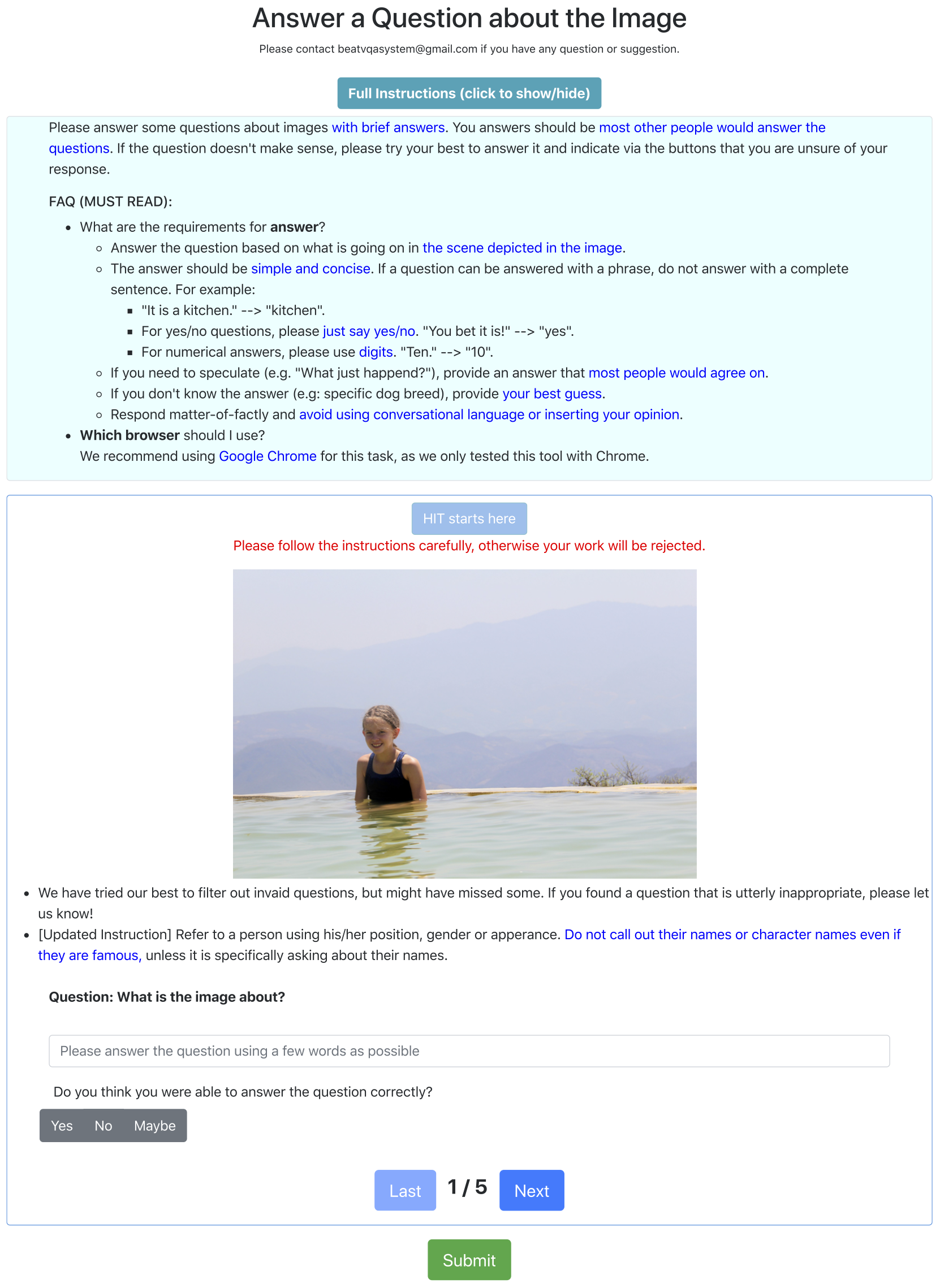}
    \caption{UI for answer collection. Given an image and a question, an annotator is asked to write a concise answer to the question, and choose a confidence level for the answer (``Yes'', ``No'', or ``Maybe'').}
    \label{fig:ans_interface}
\end{figure*}

\begin{figure*}
    \centering
    \includegraphics[height=8.5in]{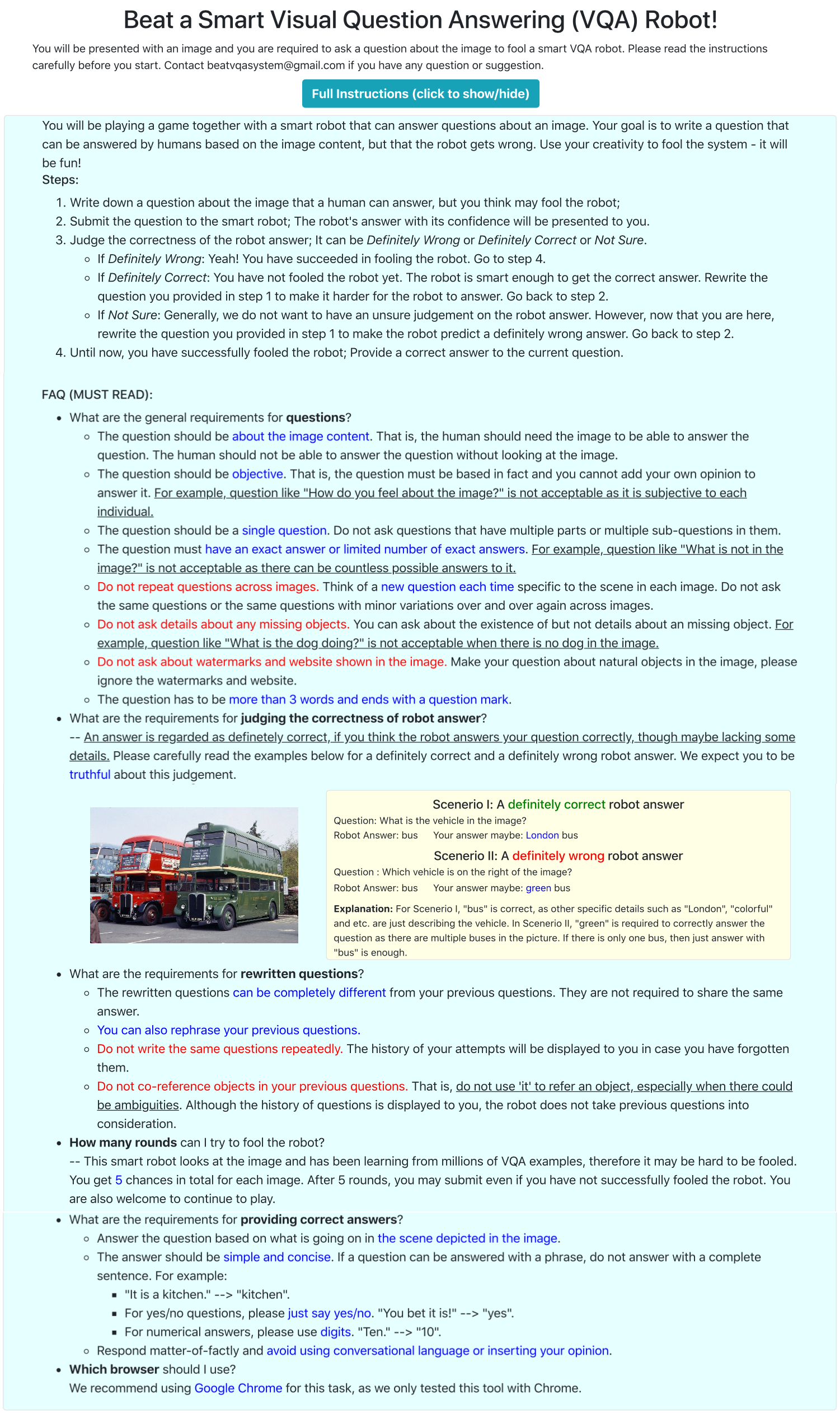}
    \caption{Full instructions for question collection.}
    \label{fig:q_instructions}
\end{figure*}

\begin{figure*}
    \centering
    \includegraphics[width=.8\textwidth]{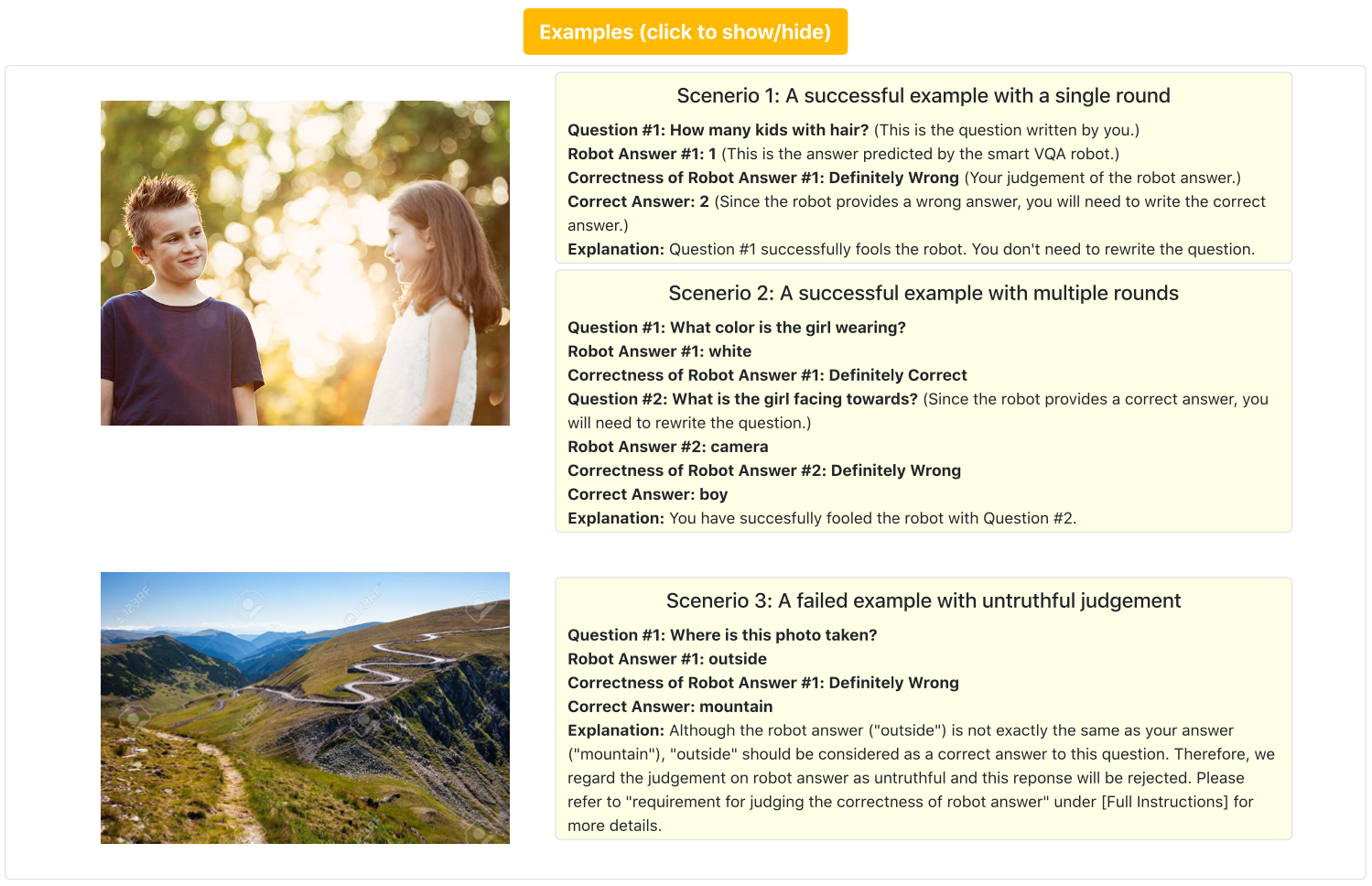}
    \caption{Examples provided to annotators for question collection.}
    \label{fig:q_examples}
\end{figure*}

Examples of the user interface are shown in Figures~\ref{fig:q_interface_step1}, ~\ref{fig:q_interface_step2} and~\ref{fig:ans_interface}. We also include full instructions and examples shown to the annotators in Figures~\ref{fig:q_instructions} and~\ref{fig:q_examples}.


\end{document}